\pdfoutput=1

\documentclass{article}

\usepackage[final]{corl_2026}           %
\makeatletter
\renewcommand{\@conferencelocation}{Austin, Texas, USA}
\makeatother

\usepackage{amsmath,amsfonts,amssymb}
\usepackage{graphicx}
\usepackage{float}     %
\usepackage{placeins}  %
\usepackage{booktabs}
\usepackage{tabularray}
\UseTblrLibrary{booktabs}
\usepackage{xcolor}
\definecolor{sdblue}{HTML}{2a78d6}
\definecolor{jrorange}{HTML}{eb6834}
\usepackage{multirow}
\usepackage{hyperref}
\usepackage{subcaption}
\usepackage{tikz}
\usetikzlibrary{positioning,arrows.meta,fit,backgrounds,calc,shapes.geometric}

\newif\ifrev
\ifdefined\showrevisions\revtrue\else\revfalse\fi
\newcommand{\rev}[1]{\ifrev\textcolor{red}{#1}\else#1\fi}

\title{To Route or to Adapt? Multi-Expert Diffusion Planners in Unseen Regions}

\author{%
  \rev{Yun Li$^{1}$\quad Ehsan Javanmardi$^{1}$\quad Simon Thompson$^{2}$\quad Jiachen Gong$^{1}$\quad Qunli Zhang$^{3}$} \\
  \rev{\bfseries Zifan Zeng$^{3}$\quad Shiming Liu$^{3}$\quad Peng Wang$^{3}$\quad Yidu Zhang$^{1}$\quad Manabu Tsukada$^{1}$} \\[2pt]
  \rev{$^{1}$The University of Tokyo \quad $^{2}$TIER IV, Inc. \quad $^{3}$Huawei} \\
  \rev{\url{https://github.com/tlab-wide/RouteOrAdapt}}
}

\begin{document}
\maketitle

\begin{abstract}
\rev{
Diffusion trajectory planners with domain-specific experts
face a practical question in unseen regions: can they
rely on selecting among existing experts, or must the
planner itself adapt? We study this question using a
planner with a shared adapter and per-domain expert
adapters. Across three real-world held-out maps, none
of eight test-time routing schemes improves the
fleet-mean lateral offset over the static default,
which uses the same expert for every scene. Even an
oracle that selects the best expert for each scene
reduces the mean offset by only $0.9$\,m. In contrast,
reward-based fine-tuning on just $20$ target-region
scenes yields substantially larger gains. On matched
evaluation slices, joint replay achieves a fleet-mean
offset of $6.9$\,m, compared with $12.4$\,m for the
static default. Comparisons under the same target-region
training budget show that target-region training drives
the improvement, while replaying training scenes limits
forgetting on the original ODDs. On nuPlan, fine-tuning
across training cities also improves open-loop offsets,
although closed-loop performance deteriorates.
Further diagnostics reveal that two test-time adaptation
baselines achieve low offsets by producing degenerate
trajectories. These results show the limited headroom
for selecting among existing experts and the greater
open-loop gains available from adapting the planner.
}
\end{abstract}

\keywords{Diffusion Planners, Out-of-Distribution Generalisation, Expert Routing, \rev{Domain Adaptation, Replay}, LoRA, Conformal Prediction}

\section{Introduction}
\label{sec:intro}

\rev{Deploying a trajectory planner in an unseen region
presents a challenge for models trained with
domain-specific experts. One approach is to select
the most suitable expert for each new scene,
reusing what the planner has already learned.
Another is to update the model using data from
the new region. The former avoids additional
training but is limited by the available experts;
the latter can change the planner's predictions
but risks degrading performance on the original
training domains. How much can expert selection
actually achieve before adaptation becomes necessary?}

\begin{figure}[t]
\centering
\includegraphics[width=\linewidth]{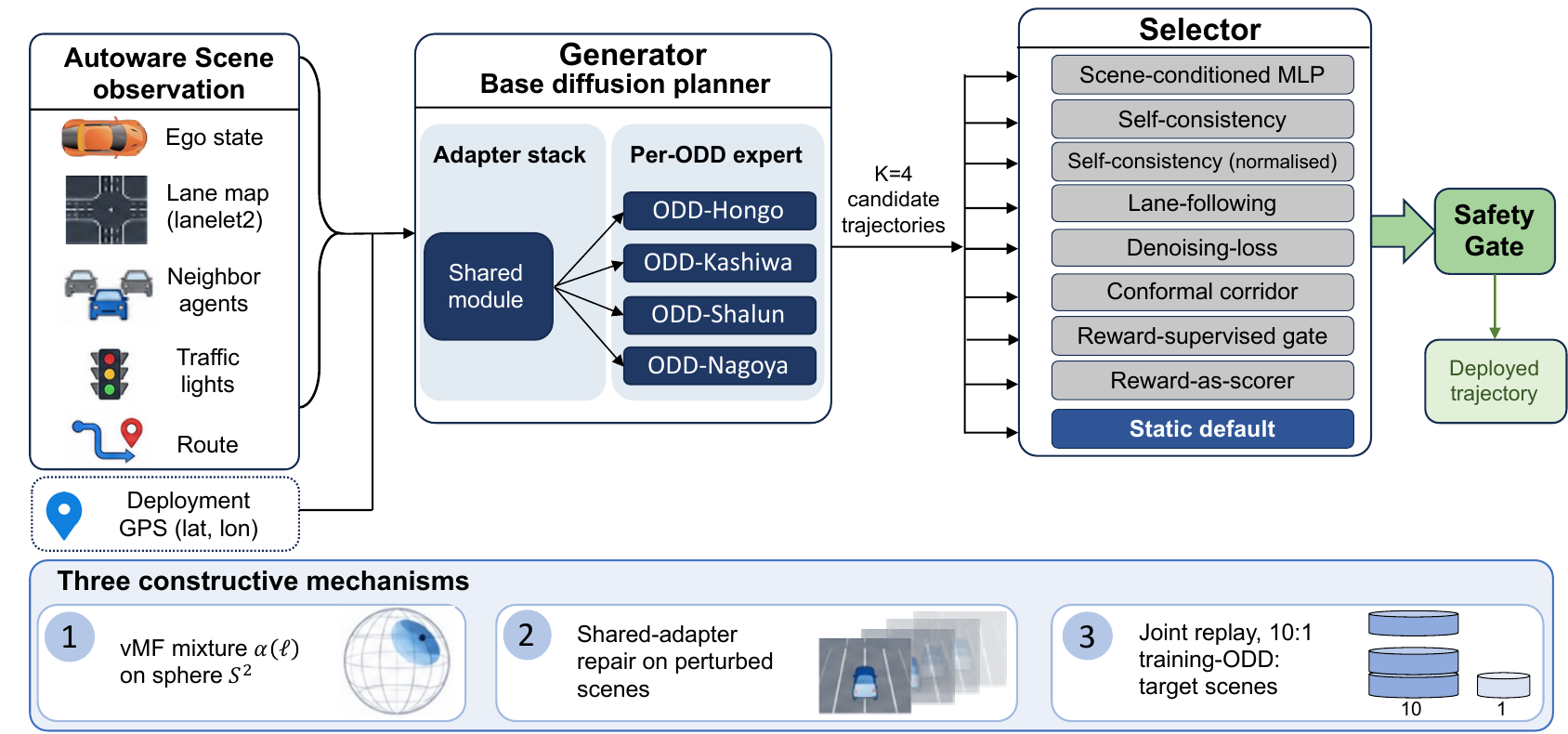}
\caption{Method overview aligned with the Autoware E2E Generator--Selector reference architecture~\citep{autoware2025e2e}. The Generator (\rev{the adapter stack}) emits $K{=}4$ candidate trajectories\rev{, and the Selector chooses one by routing or scoring; the vMF mixture and the fine-tune mechanisms of \S\ref{sec:constructive} instead change the Generator's weights;} a Safety Gate enforces the Reachability-of-Safe-State constraint.}
\label{fig:overview}
\end{figure}

\rev{We investigate this question using a reward-fine-tuned
diffusion trajectory planner~\citep{Ren2024-hy,
Zheng2024-gq} with a shared adapter and four
per-ODD expert adapters. The four training ODDs
are ODD-Hongo (the University of Tokyo's Hongo
campus), ODD-Kashiwa (suburban Kashiwa),
ODD-Shalun (a controlled test site),
and ODD-Nagoya (Nagoya University campus).
The three held-out maps are Knowledge Hub
(a research campus), Ishiodai (a suburban
district), and Morikoro (an open park), all in
Aichi Prefecture, Japan.
Within the Generator--Selector framework of the
Autoware end-to-end reference
architecture~\citep{autoware2025e2e}, we compare
methods that select among existing expert
trajectories with methods that modify the
planner's adapter weights,
as shown in Figure~\ref{fig:overview}.}

\rev{We first examine whether routing can make better
use of the existing experts. Across the three
held-out maps, none of eight routing schemes
improves the fleet-mean lateral offset over the
static default, which uses the geographically
nearest training expert (ODD-Nagoya on all three
maps). More importantly, an oracle that selects
the lowest-offset expert trajectory for each
scene improves the mean offset by only $0.9$\,m.
Even perfect selection among the four existing
trajectories therefore offers little additional
benefit. We next examine whether updating the
planner can overcome this limitation.
Fine-tuning the shared adapter on geometrically
perturbed training scenes improves performance
on the open-park map, while reward-based
fine-tuning with $20$ target-region scenes
provides substantially larger gains.
On matched evaluation slices, joint replay,
which mixes target-region and training scenes
during fine-tuning, achieves a fleet-mean
offset of $6.9$\,m, compared with $12.4$\,m
for the static default. Comparisons with
other fine-tuning methods show that
target-region training drives the improvement,
while replay primarily preserves performance
on the original training ODDs.}

\rev{We also examine the limits of these improvements.
On nuPlan, fine-tuning across three training
cities reduces open-loop lateral offset in each
held-out city but degrades closed-loop performance.
In addition, two test-time adaptation baselines
produce low lateral offsets by collapsing toward
nearly stationary or scene-independent trajectories.
These observations distinguish improvements
in open-loop trajectory metrics from improvements
in driving behaviour, which remain to be established
for the proposed adaptation methods.}

\section{Related Work}
\label{sec:related}

\paragraph{\rev{Diffusion planning and conformal prediction.}}
\rev{Our planner builds on reward-based diffusion policy
training~\citep{Ren2024-hy,Zheng2024-gq}, using a shared
adapter and per-ODD experts. Conformal prediction has
been applied to trajectory uncertainty~\citep{Lekeufack2024-wh,Lindemann2023-fi} and semantic
uncertainty in BEV planning~\citep{Ryu2025-iy}.
Selective conformal risk control combines abstention
with calibrated prediction~\citep{Xu2025-oj}.
We examine whether a conformal corridor can help
select among the trajectories of existing experts
when the deployment region differs from the
calibration domains.}

\paragraph{Geographic conditioning in driving.}
\rev{GeCo~\citep{Zhu2023-gj} uses geographic channel
attention and contrastive imitation learning to
train a policy across regions. Drive-Anywhere~\citep{Wang2023-vq} instead uses multimodal foundation
model representations to improve generalisation
to unseen environments. Our setting differs from
both: the planner and its domain-specific experts
are already trained, and we examine whether
geographic location can guide their selection
or weight mixing without retraining the backbone.}

\paragraph{\rev{Expert routing and weight merging.}}
\rev{Related approaches include online expert selection
for trajectory prediction~\citep{Tong2025-kv},
mixtures of denoising experts for robotic
manipulation~\citep{Reuss2024-zu}, and
CLIP-based LoRA retrieval for open-vocabulary
segmentation~\citep{Qorbani2025-qa}.
These methods differ in their tasks and available
supervision. We evaluate both selecting among
the trajectories generated by existing experts
and combining their adapter weights.
The latter is related to model soups and
weight-space merging~\citep{Wortsman2022-ra,
Yadav2023-fs}, although our geographic mixing
weights depend on the deployment location.}

\paragraph{\rev{Multi-task learning and domain adaptation.}}
\rev{The shared-plus-expert adapter structure is
related to multi-task learning with shared
representations~\citep{Caruana1997-zf,
Bengio2013-ad}. For adaptation, we study
fine-tuning on geometrically perturbed training
scenes, related to scene augmentation for OOD
generalisation~\citep{Yao2024-xr,Li2024-yj}
and generative simulation~\citep{Yang2025-tg,
Ren2025-sj}. We also fine-tune with a small
set of target-region scenes, using replay
of training scenes to limit forgetting~\citep{Robins1995-rehearsal,Rolnick2019-replay}.
Separately, we evaluate test-time adaptation
baselines~\citep{Wang2020-io,Niu2022-zu}
and analyse their degenerate predictions
in App.~\ref{app:tta} and
App.~\ref{app:hypernet}.}

\section{Method}
\label{sec:method}

\paragraph{Generator--Selector decomposition.}\label{subsec:generator_selector}
\rev{We follow the Generator--Selector framework of
the Autoware end-to-end reference
architecture~\citep{autoware2025e2e},
as shown in Figure~\ref{fig:overview}.
The Generator consists of a diffusion planner~\citep{Zheng2024-gq} with a shared low-rank
adapter (LoRA~\citep{Hu2022-lora}) and four
per-ODD expert adapters of comparable rank.
For each scene, we activate the shared adapter
together with each expert in turn, producing
$K=4$ candidate trajectories.
The Selector ranks these candidates and
returns a single trajectory.}

\rev{We evaluate eight routing schemes that choose among
or weight the existing candidates, alongside three
mechanisms that modify the adapter weights
before generation. The conformal routing scheme
also uses geographic weight mixing as a fallback,
as described in \S\ref{sec:gps_mix}.
The adapter ranks and optimisation settings are
inherited from the training of the adapter stack
(App.~\ref{app:hp}).
The per-ODD structure supports independent
updates and rollbacks for regions with different
data volumes and maintenance schedules.
Our experiments do not cover monolithic planners
trained on pooled data.}

\paragraph{\rev{Eight routing schemes.}}
\label{subsec:routing}
\label{subsec:conformal}
\label{subsec:backbone}
\rev{We evaluate eight schemes grouped by the
information used to choose an expert.}

\rev{\emph{Feature and geometry.}
(i) A two-hidden-layer MLP takes eight groups
of scene features: lane count, mean curvature,
mean lane length, lateral and longitudinal lane
spread, neighbour speed, route-lane count, and
ego state.
(ii) Lane-following scores each candidate by
its perpendicular distance to the nearest
route-lane centreline.}

\rev{\emph{Consistency and uncertainty.}
(iii) Self-consistency scores the agreement
among five trajectories generated by each expert,
using mean pairwise trajectory distance.
(iv) A magnitude-normalised variant uses the
same samples.
(v) A split-conformal corridor uses a multi-noise
nonconformity score calibrated on training scenes
(App.~\ref{app:conformal}).}

\rev{\emph{Loss and reward.}
(vi) Denoising-loss scoring ranks experts using
their denoising losses~\citep{Lee2025-bg}.
(vii) A reward-supervised dynamic gate is trained
directly on realised lateral offset
(\S\ref{sec:routing_fails}).
(viii) Reward-as-scorer selects
$\arg\max_e R(\tau_e)$ using the planner's
training reward.}

\rev{We compare each scheme with the static default
on the scenes where it is evaluated.}

\begin{figure}[t]
\centering
\includegraphics[width=\linewidth,trim=0 46 0 0,clip]{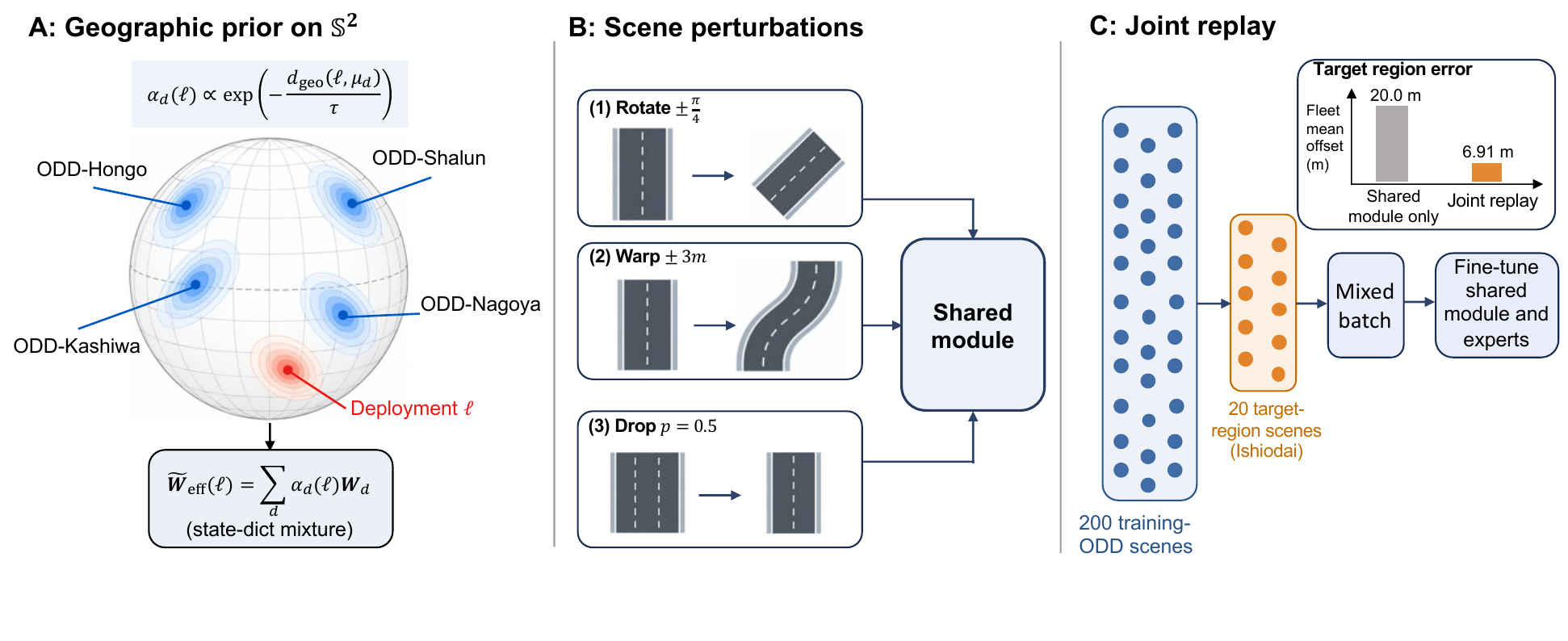}
\caption{Three constructive mechanisms. \rev{(A) vMF mixture: experts weighted by distance to the deployment location (\S\ref{sec:gps_mix}). (B) Shared-adapter repair: only the shared module is fine-tuned, on perturbed training scenes (\S\ref{sec:park_repair}). (C) Joint replay: the whole adapter stack is fine-tuned on training and target-region scenes; inset: target-region error (\S\ref{sec:joint_replay_frontier}).}}
\label{fig:methods}
\end{figure}

\paragraph{\rev{Three constructive mechanisms.}}
\label{subsec:mechs}
\rev{We also investigate three ways to modify the
planner's adapters, shown in Figure~\ref{fig:methods}.
First, geographic weight mixing (the vMF mixture)
combines the existing expert weights according to
their distances from the deployment location
(\S\ref{sec:gps_mix}).
Second, shared-adapter repair fine-tunes the
shared module on training scenes augmented
with lanelet rotations, centreline warps,
and boundary removal
(\S\ref{sec:park_repair}).
Third, joint replay fine-tunes the full adapter
stack using target-region scenes together
with $200$ training scenes
(\S\ref{sec:joint_replay_frontier}).
The latter two methods update model parameters,
whereas geographic mixing combines existing
parameters without further optimisation.}

\paragraph{Reward and safety gate.}
\label{subsec:reward}
\rev{We use the trajectory-level reward inherited
from the training of the adapter stack, with terms
for forward progress, route alignment, comfort,
and safety.
The safety term draws on the
Reachability-of-Safe-State requirement discussed
in the Autoware reference
architecture~\citep{autoware2025e2e}.
The full reward, its weights, and the
braking-distance parameters are provided
in App.~\ref{app:hp}.}

\rev{A Safety Gate downstream of the Selector rejects
trajectories that violate hard limits on lateral
offset, braking distance, or forward progress.
All lateral-offset results in this paper are
measured on the Selector's output before
the Safety Gate.}

\paragraph{\rev{Terms and evaluation protocol.}}
\label{subsec:eval}
\rev{We evaluate planned trajectories over $8$\,s
at $10$\,Hz. For a trajectory
$\tau=\{(x_t,y_t)\}_{t=1}^{80}$ expressed in
the ego frame aligned with the start lanelet,
the lateral offset is the maximum absolute
lateral coordinate. The fleet mean gives
equal weight to the three held-out maps:
\begin{equation}
o(\tau)=\max_t |y_t|,
\qquad
\bar{o}=
\frac{1}{3}\sum_m
\frac{1}{|\mathcal{S}_m|}
\sum_{s\in\mathcal{S}_m}o(\tau_s),
\label{eq:offset}
\end{equation}
where $\mathcal{S}_m$ is the evaluation scene
set for map $m$. This metric measures lateral
displacement relative to the initial lanelet
axis, rather than distance to the route
centreline or a recorded trajectory.}

\rev{The \emph{static default} always activates
the expert of the geographically nearest
training ODD. For all three held-out maps,
this is ODD-Nagoya, which also has the
lowest offset among the four fixed experts.}

\rev{We define \emph{oracle headroom} as the
improvement obtained by selecting, for each
scene, the lowest-offset trajectory among
the four experts:
\begin{equation}
H_{\mathrm{oracle}}=
\frac{1}{N}\sum_{s=1}^{N}
\left[
o(\tau_s^{\mathrm{static}})
-
\min_e o(\tau_s^e)
\right].
\label{eq:oracle_headroom}
\end{equation}
This measures the potential gain from selecting
among the four existing candidate trajectories;
it does not bound the performance of weight
mixing or fine-tuning.}

\rev{We use two evaluation pools.
\emph{Pool~A} contains $24$ scenes used for the
routing experiments.
\emph{Pool~B} is used for fine-tuning evaluation
and consists, for each training seed, of scenes
drawn from lanelets disjoint from that seed's
training scenes.
Table~\ref{tab:setup} specifies the evaluation
pool for each experiment. Results from
different pools are reported separately.}

\begin{table}[t]
\centering\footnotesize
\caption{\rev{Model, training and evaluation pool of each experiment (pools defined in \S\ref{subsec:eval}).}}
\label{tab:setup}
\renewcommand{\arraystretch}{1.12}
\begin{tabular*}{\linewidth}{@{\extracolsep{\fill}}llll@{}}
\toprule
\rev{Experiment} & \rev{Model evaluated} & \rev{Training} & \rev{Evaluation} \\
\midrule
\rev{Routing schemes (\S\ref{sec:routing_fails})}           & \rev{adapter stack}                & \rev{selector only}                  & \rev{Pool~A ($24$)} \\
\rev{Oracle headroom (\S\ref{sec:routing_fails})}           & \rev{per-scene oracle}             & \rev{none}                           & \rev{Pool~A; $104$ scenes} \\
\addlinespace[3pt]
\rev{vMF mixture (\S\ref{sec:gps_mix})}                     & \rev{average of experts}           & \rev{closed-form weights}            & \rev{Pool~A ($24$)} \\
\rev{Shared-adapter repair (\S\ref{sec:park_repair})}       & \rev{shared module}                & \rev{$500$ steps, perturbed scenes}  & \rev{Pool~A ($24$)} \\
\rev{Joint replay, baselines (\S\ref{sec:joint_replay_frontier})} & \rev{adapter stack}          & \rev{$500$ steps, $20$ target scenes} & \rev{Pool~B, $8$ seeds} \\
\addlinespace[3pt]
\rev{nuPlan cross-city (\S\ref{sec:realvehicle})}           & \rev{rank-$4$ shared LoRA}         & \rev{$500$ steps on $3$ cities}      & \rev{nuPlan, open loop} \\
\rev{Real vehicle, DPO (\S\ref{sec:realvehicle})}           & \rev{other adapters}               & \rev{reward or DPO}                  & \rev{ODD-Kashiwa course} \\
\bottomrule
\end{tabular*}
\end{table}

\section{\rev{Limited Headroom for Expert Routing}}
\label{sec:routing_fails}
\label{sec:routing}
\label{sec:result1}
\label{sec:result2}
\label{subsec:mi}

\rev{
We begin by measuring how much could be gained from
selecting among the four existing experts. For each
scene, an oracle chooses the expert trajectory with
the lowest lateral offset. This gives an upper
bound on the improvement available to any selector
restricted to those four trajectories, though not
to methods that combine expert weights or update
the planner.
}

\rev{
On Pool~A, the oracle improves the mean
offset over the static default by only $0.9$\,m.
The gain remains small on a larger $104$-scene
pool with new start lanelets ($1.3$\,m) and on
the disjoint evaluation slices of Pool~B
($1.1$\,m). Thus, most of the offset cannot
be reduced by choosing a different existing
expert for each scene.
For comparison, joint replay
(\S\ref{sec:joint_replay_frontier}) reduces
the mean offset on Pool~A by $9.1$\,m.
Seven of these scenes are among joint replay's
training scenes; on the remaining $17$ scenes,
the improvement is $12.7$\,m ($p=0.0008$).
}

\begin{figure}[t]
\centering
\includegraphics[width=0.97\linewidth]{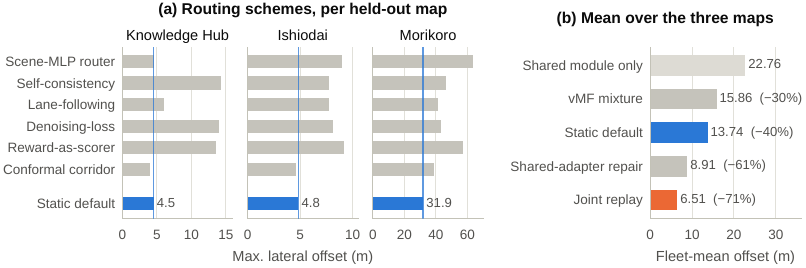}
\caption{\rev{(a) Lateral offset of the routing schemes (gray) and the static default (blue) on each held-out map; two schemes are omitted because the normalised self-consistency equals the vanilla one and the reward-supervised gate ran on Morikoro only. (b) Fleet mean of the three constructive mechanisms and its change relative to the shared module only. Pool~A; joint replay excludes its $7$ Ishiodai training scenes.}}
\label{fig:results}
\end{figure}

\rev{
We next evaluate eight routing schemes on Pool~A,
reported in Table~\ref{tab:routing} and
Figure~\ref{fig:results}a.
The static default uses the geographically
nearest expert, ODD-Nagoya, which is also the
best fixed expert on all three held-out maps.
It achieves a fleet-mean offset of $13.74$\,m.
The conformal corridor slightly improves on
this baseline on Knowledge Hub and Ishiodai,
but performs worse on Morikoro, leaving its
fleet mean higher than the static default.
None of the schemes evaluated across all three
maps improves the fleet mean.
Paired Wilcoxon tests find no significant
improvement, while six schemes perform
significantly worse ($p\leq 0.012$; all six
remain significant after Holm correction).
Morikoro contributes the largest offsets.
Its open-park layout differs from the training
ODDs, all of which contain paired opposing
lanes (App.~\ref{app:topologies}).
}

\begin{table}[t]
\centering\footnotesize
\caption{Eight routing schemes versus the static \rev{default: lateral offset (m) on Pool~A, mean $\pm$ std over scenes, and the fleet mean. Six of the eight are significantly worse than the default under paired per-scene tests ($p \leq 0.012$), none significantly better. The reward-supervised gate was run on Morikoro only ($\pm$ over $50$ seeds).}}
\label{tab:routing}\label{tab:perfix}
\renewcommand{\arraystretch}{1.12}
\begin{tblr}{colspec={llllllll}, colsep=7pt, column{1}={leftsep=0pt}, column{8}={rightsep=0pt}, column{2,4,6}={halign=r,rightsep=0pt}, column{3,5,7}={halign=l,leftsep=0pt}, column{8}={halign=r}, row{10}={bg=sdblue!10}}
\toprule
\rev{Routing scheme} & \SetCell[c=2]{c} Knowledge Hub & & \SetCell[c=2]{c} Ishiodai & & \SetCell[c=2]{c} Morikoro & & Fleet \\
\midrule
Scene-conditioned MLP & \rev{4.5} & \rev{\,$\pm$\,0.8} & \rev{9.0} & \rev{\,$\pm$\,4.2} & \rev{63.3} & \rev{\,$\pm$\,58.0} & \rev{25.61} \\
Self-consistency & 14.3 & \,$\pm$\,4.1 & 7.7 & \,$\pm$\,4.6 & 46.3 & \,$\pm$\,35.3 & \rev{22.76} \\
Self-consistency (normalised) & 14.3 & \,$\pm$\,4.1 & 7.7 & \,$\pm$\,4.6 & 46.3 & \,$\pm$\,35.3 & \rev{22.76} \\
Lane-following & 6.0 & \,$\pm$\,2.1 & \rev{7.8} & \rev{\,$\pm$\,4.0} & \rev{41.1} & \rev{\,$\pm$\,46.9} & \rev{18.29} \\
Denoising-loss & \rev{14.0} & \rev{\,$\pm$\,4.0} & \rev{8.1} & \rev{\,$\pm$\,4.3} & \rev{42.9} & \rev{\,$\pm$\,33.7} & \rev{21.67} \\
Conformal corridor ($T{=}15$) & \textbf{4.0} & \,$\pm$\,\textbf{0.1} & \textbf{4.6} & \,$\pm$\,\textbf{0.7} & 38.9 & \,$\pm$\,37.1 & \rev{15.86} \\
Reward-supervised gate & \SetCell[c=2]{c} n/a & & \SetCell[c=2]{c} n/a & & 32.3 & \,$\pm$\,0.6 & \rev{n/a} \\
Reward-as-scorer & \rev{13.6} & \rev{\,$\pm$\,3.6} & \rev{9.2} & \rev{\,$\pm$\,4.7} & \rev{56.9} & \rev{\,$\pm$\,51.8} & \rev{26.57} \\
\midrule
\textbf{Static default} (ODD-Nagoya) & \rev{4.5} & \rev{\,$\pm$\,0.8} & \rev{4.8} & \rev{\,$\pm$\,0.8} & \rev{\textbf{31.9}} & \rev{\textbf{\,$\pm$\,32.1}} & \textbf{13.74} \\
\bottomrule
\end{tblr}
\end{table}

\rev{
The routing schemes encounter different problems
on the held-out maps.
The scene-conditioned MLP separates the training
ODDs perfectly but assigns most held-out scenes
to ODD-Hongo, resulting in large offsets.
Self-consistency favours small-magnitude
trajectories and effectively selects the shared
module alone.
The denoising loss has little correlation with
realised offset, while the planner's own reward
frequently selects ODD-Shalun instead of the
lowest-offset expert.
The reward-supervised gate converges almost
entirely to ODD-Nagoya, recovering approximately
the static default.
The conformal corridor abstains on every scene
at $T=5$\,m; at $T=15$\,m, its output reduces
to the geographic weight mixture of
\S\ref{sec:gps_mix}.
Further results are provided in
App.~\ref{app:routing_details}.
}

\rev{
We also test whether scene features can predict
which expert will have the lowest offset.
Leave-one-out classifiers using either
hand-crafted scene features or the planner's
encoder embeddings do not outperform the
majority-class baseline on the $24$- or
$104$-scene pools (permutation $p\geq 0.29$).
A richer $38$-dimensional feature set separates
the training ODDs perfectly but does not
improve selection on the held-out maps
(App.~\ref{app:feature_ablation}).
A sum of per-coordinate mutual-information
estimates~\citep{Kraskov2004-ksg} between the
encoder embedding and the oracle label is at
most suggestive ($p{=}0.088$ with $24$ scenes
in $256$ dimensions).
The conformal predictor faces a separate
problem: its training and held-out residuals
fail an exchangeability test ($p=0.0035$;
App.~\ref{app:conformal}).
}

\rev{
These results show that the tested representations
and routing methods do not reliably identify
better experts under cross-region shift.
Together with the small oracle headroom, they
suggest that more effective selection among the
four existing experts would offer limited
benefit on these maps.
}

\section{\rev{Beyond Expert Selection: Weight Mixing and Adaptation}}
\label{sec:constructive}

\rev{We next examine three alternatives to selecting among
existing expert trajectories. Geographic weight mixing
combines pretrained experts without further training;
shared-adapter repair uses geometrically perturbed
training scenes; and joint replay introduces a small
number of target-region scenes. Only the last method
requires target-region training data.
Figure~\ref{fig:results}b compares their results.}

\subsection{\rev{Geographic Weight Mixing}}
\label{sec:gps_mix}\label{subsec:gps_mix}

\rev{We first test whether geographic proximity can
guide the combination of existing experts.
Let $d_{\mathrm{geo}}(\ell,\mu_d)$ denote the
great-circle distance between deployment location
$\ell$ and the centre $\mu_d$ of training ODD $d$.
We compute a weighted average of the merged
expert weights $W_d$:}
\begin{equation}
\widetilde{W}(\ell)
= \sum_d \alpha_d(\ell)\,W_d,
\qquad
\alpha_d(\ell)
= \frac{\exp(-d_{\mathrm{geo}}(\ell,\mu_d)/\tau)}
{\sum_j \exp(-d_{\mathrm{geo}}(\ell,\mu_j)/\tau)}.
\label{eq:vmf}
\end{equation}
\rev{The temperature $\tau$ controls the contribution
of distant experts. We evaluate
$\tau\in\{1,10,50,200\}$\,km; as $\tau$ approaches
zero, the mixture approaches the nearest-expert
static default (App.~\ref{app:pacbayes}).}

\rev{At $\tau=200$\,km, the mixture slightly improves
Knowledge Hub and Ishiodai but increases the
Morikoro offset from $31.9$ to $38.9$\,m.
Its fleet mean is $15.86$\,m, compared with
$13.74$\,m for the static default
(Table~\ref{tab:gps}).
Learning the mixture coefficients directly
from $20$ target-region scenes performs worse
(App.~\ref{app:learned_coef}).
Neither approach improves the fleet mean over
the static default.}

\subsection{\rev{Shared-Adapter Repair with Geometric Perturbations}}
\label{sec:park_repair}

\rev{We next fine-tune the shared adapter without
using target-region training scenes.
For $500$ steps, we train on scenes from the
original ODDs with three geometric perturbations:
per-lanelet SE(2) rotations of up to $45^\circ$,
cubic Bezier centreline warps of up to $3$\,m,
and random removal of one lane boundary with
probability $0.5$.
These perturbations introduce variations in
lane orientation, curvature, and boundary
structure that are relevant to the open-park map.}

\rev{The repaired shared adapter more than halves
Morikoro's offset relative to the original
shared adapter and also improves the other two
maps. When combined with the ODD-Kashiwa expert,
it achieves $11.1$\,m on Morikoro, compared
with $31.9$\,m for the static default.
Ablations across three seeds show that lanelet
rotation contributes the largest improvement
(App.~\ref{app:park_repair}).}

\subsection{\rev{Joint Replay with Target-Region Scenes}}
\label{sec:joint_replay_frontier}

\rev{We then fine-tune the full adapter stack $\theta$
using target-region scenes together with scenes
from the original training ODDs.
The training set contains $20$ target-region
scenes from Ishiodai ($\mathcal{D}_{\mathrm{tgt}}$)
and $200$ training scenes
($\mathcal{D}_{\mathrm{src}}$), with $50$ from
each training ODD.
Replaying training scenes follows the rehearsal
approach to continual learning~\citep{Robins1995-rehearsal,Rolnick2019-replay}.
We optimise the trajectory-level reward for
$500$ steps:}
\begin{equation}
\max_{\theta}
\sum_{x\in\mathcal{D}_{\mathrm{src}}
\cup\mathcal{D}_{\mathrm{tgt}}}
R_{c(x)}\bigl(\tau_\theta(x);x\bigr).
\label{eq:joint_replay}
\end{equation}
\rev{Here $\tau_\theta(x)$ is the trajectory
generated by a differentiable DPM-Solver$++$
rollout, and $R_{c(x)}$ uses the reward
weights assigned to the scene's ODD.
Training scenes activate their respective
experts, while target-region scenes use
the shared adapter alone with ODD-Nagoya
reward weights. The reward is computed
from scene maps and routes, without
requiring recorded driving trajectories.}

\rev{On Pool~B, joint replay achieves a fleet-mean
offset of $6.91\pm2.16$\,m across eight seeds,
compared with $12.4$\,m for the static default
(Table~\ref{tab:matched_baselines}).\footnote{A single checkpoint reaches $4.26$\,m and the three-seed sweep of Figure~\ref{fig:sample_complexity} gives $6.17{\pm}1.23$\,m at $n{=}20$; we quote the eight-seed number throughout.}
Because all target-region training scenes
come from Ishiodai, improvements on Knowledge
Hub and Morikoro reflect transfer to the
other two held-out maps.
The sample-complexity results in
Figure~\ref{fig:sample_complexity} also show
reduced variation across seeds as the target
training set grows to $20$ scenes.}

\paragraph{\rev{Target-region training and replay.}}
\rev{To separate the effects of target-region
training and replay, we compare methods using
the same $20$ target scenes, optimisation steps,
LoRA ranks, and learning rate
(Table~\ref{tab:matched_baselines}).
All methods trained on target-region scenes
achieve offsets of approximately $4$--$8$\,m.
Target-only fine-tuning with
$L_2$-SP~\citep{Li2018-l2sp} gives the
lowest target offset ($4.04$\,m), although
its $95$th-percentile jerk is $27\%$ higher
than that of joint replay.}

\rev{The main difference is performance retention
on the training ODDs. Fine-tuning without
replay or EWC increases their summed offset by
$17$--$38$\,m, including when $L_2$-SP
regularisation is used.
EWC~\citep{Kirkpatrick2017-ewc} reduces
forgetting at the cost of higher target
offsets. Joint replay limits the summed
regression to $0.9\pm1.4$\,m while retaining
a target offset of $6.91\pm2.16$\,m.
Prioritised replay shows no clear advantage
over uniform replay.
These comparisons indicate that target-region
training accounts for the offset reduction,
whereas replay mainly preserves performance
on the original training ODDs.}

\begin{table}[t]
\begin{minipage}[t]{0.555\linewidth}
\centering\footnotesize
\captionof{table}{\rev{Baselines with the same $20$ target scenes, $500$ steps, LoRA ranks and learning rate; mean $\pm$ std over eight seeds, each on its own held-out slice (Pool~B). Forgetting is the summed increase in lateral offset over the four training ODDs.}}
\label{tab:matched_baselines}
\setlength{\tabcolsep}{3pt}
\renewcommand{\arraystretch}{1.12}
\begin{tblr}{colspec={lllll}, colsep=3.5pt, column{1}={leftsep=0pt}, column{5}={rightsep=0pt}, column{2,4}={halign=r,rightsep=0pt}, column{3,5}={halign=l,leftsep=0pt}, row{3}={bg=sdblue!10}, row{10}={bg=jrorange!12}, row{4,9}={abovesep=4pt}}
\toprule
\rev{Method} & \SetCell[c=2]{c} \rev{Target [m] $\downarrow$} & & \SetCell[c=2]{c} \rev{Forgetting [m] $\downarrow$} & \\
\midrule
\rev{Shared module only} & \rev{20.0} & \rev{\,$\pm$\,2.4} & \SetCell[c=2]{c} \rev{0} &  \\
\rev{Static default} & \rev{12.4} & \rev{\,$\pm$\,3.0} & \SetCell[c=2]{c} \rev{0} &  \\
\rev{Fine-tune best fixed expert} & \rev{4.69} & \rev{\,$\pm$\,1.90} & \rev{$+$22.6} & \rev{\,$\pm$\,2.9} \\
\rev{Target-only LoRA} & \rev{5.34} & \rev{\,$\pm$\,4.46} & \rev{$+$37.4} & \rev{\,$\pm$\,11.9} \\
\rev{Target-only $+$ $L_2$-SP} & \rev{\textbf{4.04}} & \rev{\,$\pm$\,\textbf{0.77}} & \rev{$+$37.8} & \rev{\,$\pm$\,13.2} \\
\rev{Adapter from scratch} & \rev{6.03} & \rev{\,$\pm$\,2.64} & \rev{$+$17.3} & \rev{\,$\pm$\,23.4} \\
\rev{EWC ($\lambda{=}10^{2}$, $3$ seeds)} & \rev{8.20} & \rev{\,$\pm$\,0.83} & \rev{$+$4.7} & \rev{\,$\pm$\,1.3} \\
\rev{Prioritised replay} & \rev{6.78} & \rev{\,$\pm$\,1.33} & \rev{$+$1.7} & \rev{\,$\pm$\,0.8} \\
\rev{\textbf{Joint replay (ours)}} & \rev{6.91} & \rev{\,$\pm$\,2.16} & \rev{\textbf{$+$0.9}} & \rev{\textbf{\,$\pm$\,1.4}} \\
\bottomrule
\end{tblr}
\end{minipage}\hfill
\begin{minipage}[t]{0.415\linewidth}
\centering
\vspace{0pt}
\includegraphics[width=\linewidth]{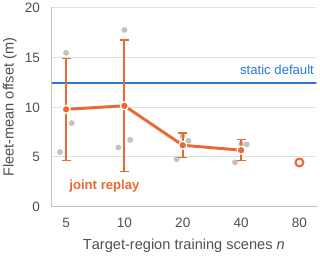}
\setlength{\abovecaptionskip}{7pt}%
\captionof{figure}{\rev{Joint replay with $n$ target-region training scenes: mean $\pm$ std over three seeds (gray dots: single seeds), each on its own held-out slice. Blue: static default on such slices; hollow marker: one seed at $n{=}80$ (App.~\ref{app:sample_complexity_multiseed}).}}
\label{fig:sample_complexity}
\end{minipage}
\end{table}

\paragraph{\rev{Additional trajectory diagnostics.}}
\rev{The target-region scenes contain no
neighbouring agents, so time-to-collision
and collision rates cannot be evaluated.
We instead report additional trajectory
diagnostics in App.~\ref{app:joint_replay_safety}.
Jerk and lateral acceleration are calculated
from unsmoothed finite differences of the
raw $10$\,Hz trajectories, so only comparisons
between models are meaningful.}

\rev{Compared with the static default, joint replay
reduces the $95$th-percentile jerk by $26\%$
and increases forward progress over $8$\,s
from $164$ to $187$\,m. Lateral acceleration
changes little, and large lateral excursions
relative to the initial lanelet remain.
The from-scratch variant of joint replay reduces
lateral acceleration and the frequency of
excursions beyond the straight $\pm3$\,m band,
but substantially degrades performance on the
training ODDs
(App.~\ref{app:joint_replay_frontier}).}

\subsection{\rev{Variance-Collapse Diagnostics for Test-Time Adaptation}}
\label{sec:diagnosed_failures}
\label{sec:variance_forensics}

\rev{Low lateral offset alone does not establish
that a trajectory is useful.
We examine two test-time adaptation methods
that achieve low offsets but produce
degenerate trajectories: TENT~\citep{Wang2020-io},
which updates normalisation parameters during
inference, and a hypernetwork that predicts
LayerNorm offsets from geographic location.}

\rev{We use two diagnostics: the variance
$\sigma_K^2$ across $K$ samples from the
same scene and the mean pairwise $L_2$
distance between trajectories from different
scenes. Starting from the static default,
TENT achieves an offset of $3.37$\,m,
but its cross-scene trajectory distance
falls from over $100$\,m to approximately
$10$\,m, consistent with similar predictions
across different scenes.
The hypernetwork reduces $\sigma_K^2$
by two orders of magnitude and produces
only $12$\,m of forward progress over $8$\,s.
These cases illustrate how low lateral offset
can accompany a loss of trajectory diversity
or forward motion
(App.~\ref{app:tta},
App.~\ref{app:hypernet}).}

\section{\rev{Cross-City Evaluation and Real-Vehicle Runs}}
\label{sec:realvehicle}

\paragraph{\rev{Cross-city evaluation on nuPlan.}}
\rev{
We evaluate cross-city transfer on nuPlan using four
leave-one-city-out splits. For each split, we fine-tune
a rank-$4$ shared LoRA on the other three cities using
the planner's diffusion training loss. Across three
seeds and using the same lateral-offset metric as on
the held-out maps, fine-tuning reduces the open-loop
lateral offset on Singapore from $8.64$ to
$5.60{\pm}0.78$\,m, by $35\%$, despite the difference
in traffic direction. Improvements over the base
planner are also observed in the other held-out
cities (App.~\ref{app:nuplan}).
Unlike the target-region experiments, this evaluation
uses no data from the held-out city and measures
transfer from fine-tuning on the training cities.
}

\paragraph{\rev{Real-vehicle runs of the base planner.}}
\rev{
The base diffusion planner has also been tested
on a physical vehicle with two adapter configurations:
a production DPO LoRA trained on human preferences
and a reward-trained shared-plus-expert adapter.
For the latter, we conducted two runs on the
ODD-Kashiwa course with the ODD-Kashiwa expert active.
The first completed the course, with a
comfort-violation rate of $47.7\%$ under our
proxy, compared with $37.7\%$ for a human driver.
In the second run, the vehicle travelled $72$\,m
at speeds no greater than $0.23$\,m/s before
stopping.
These runs document the behaviour of the base
planner on a physical vehicle; the joint-replay
adapters have not yet been tested on the vehicle.
Additional results for the DPO adapter are given
in App.~\ref{app:dpo_v3}.
}

\section{Discussion, Limitations, and Conclusion}
\label{sec:discussion}

\rev{
Our experiments show that selecting among the
four existing experts offers limited improvement
on the held-out maps, even with per-scene oracle
selection. Updating the planner with a small
amount of target-region data produces much
larger reductions in open-loop lateral offset.
Under matched training conditions, replaying
training scenes mainly helps retain performance
on the original ODDs rather than accounting
for the target-region gains.
}

\paragraph{\rev{Geographic proximity.}}
\rev{
On the three primary held-out maps, ODD-Nagoya
is both the nearest training ODD and the best
fixed expert. We examine two additional regions
to separate these effects. In a suburban area
adjacent to ODD-Kashiwa but outside its training
footprint, the nearest expert performs
$1.4$\,m worse than ODD-Nagoya
($p=0.001$, $n=30$).
In a dense urban region nearest to ODD-Hongo,
the ODD-Hongo expert performs worst among the four
(App.~\ref{app:switching}).
Geographic proximity therefore does not
consistently identify the best expert.
The success of the static default on our
primary maps appears to reflect the strength
of ODD-Nagoya rather than its proximity alone.
}

\paragraph{Limitations.}
\rev{
The main limitation is that all results on the
held-out maps are open-loop. In nuPlan's reactive
closed-loop simulator, the fine-tuned models score
below the unadapted planner in all four held-out
cities, with significant differences in three.
For Singapore, the simulator score decreases
from $0.858$ to $0.598$, and drivable-area
compliance from $0.97$ to $0.76$.
The joint-replay adapters have not been
evaluated on a physical vehicle.
}

\rev{
The primary held-out evaluation is also limited
in scale, covering $4$--$10$ road lanelets
per map. We extend the oracle and feature-classifier
analyses to $104$ scenes, but the full eight-method
routing comparison remains on the smaller pool.
Our findings therefore concern the tested
experts, routing methods, and representations.
Their applicability to monolithic planners
trained on pooled data~\citep{Zheng2024-gq}
remains unknown.
}

\rev{
The maps are proprietary. We release the
model weights, scene splits, evaluation pipeline,
and per-scene results through the project
repository.
}

\paragraph{\rev{Conclusion.}}
\rev{
Overall, the results favour updating the planner
over selecting among its existing experts when
the objective is to reduce open-loop lateral
offset in unseen regions. Whether these gains
translate into better closed-loop driving remains
an open question.
}

\acknowledgments{\rev{This work was supported by JSPS KAKENHI Grant Number JP26KJ0799 and by TIER IV, Inc.}}

\bibliography{refs}

\appendix

\section*{\rev{Overview of the Appendix}}
\rev{This appendix gives the details behind the main text. Table~\ref{tab:app_overview} lists each appendix section next to the main-text section it supports. Appendix~\ref{app:dpo_v3} reports the real-vehicle runs and a separate study of DPO adapters on the same base planner; it is context for \S\ref{sec:realvehicle} and does not evaluate the adaptation methods proposed in this paper.}

\begin{table}[h]
\centering\footnotesize\setlength{\tabcolsep}{4pt}
\caption{\rev{Guide to the appendix, ordered by the main-text section each part supports.}}
\label{tab:app_overview}
\begin{tabular}{llcl}
\toprule
\rev{Main text} & \rev{Appendix} & \rev{Page} & \rev{Content} \\
\midrule
\rev{\S\ref{sec:method}} & \rev{\ref{app:hp}} & \rev{\pageref{app:hp}} & \rev{Hyperparameters, training recipes, train/eval split tooling} \\
\midrule
\rev{\S\ref{sec:routing_fails}} & \rev{\ref{app:topologies}} & \rev{\pageref{app:topologies}} & \rev{Held-out target-region topologies} \\
 & \rev{\ref{app:routing_details}} & \rev{\pageref{app:routing_details}} & \rev{Why each routing scheme fails; oracle details} \\
 & \rev{\ref{app:feature_ablation}} & \rev{\pageref{app:feature_ablation}} & \rev{Routing feature ablations} \\
 & \rev{\ref{app:conformal}} & \rev{\pageref{app:conformal}} & \rev{Conformal calibration} \\
 & \rev{\ref{app:switching}} & \rev{\pageref{app:switching}} & \rev{Regions where the nearest expert is not ODD-Nagoya (also \S\ref{sec:discussion})} \\
\midrule
\rev{\S\ref{sec:gps_mix}} & \rev{\ref{app:pacbayes}} & \rev{\pageref{app:pacbayes}} & \rev{Derivation and weights of the vMF mixture} \\
 & \rev{\ref{app:weight_avg}} & \rev{\pageref{app:weight_avg}} & \rev{Weight averaging and learned mixing coefficients} \\
 & \rev{\ref{app:skyline}} & \rev{\pageref{app:skyline}} & \rev{Capacity test with a rank-$128$ LoRA} \\
\midrule
\rev{\S\ref{sec:park_repair}} & \rev{\ref{app:park_repair}} & \rev{\pageref{app:park_repair}} & \rev{Shared-adapter repair: compositions and ablation} \\
\midrule
\rev{\S\ref{sec:joint_replay_frontier}} & \rev{\ref{app:joint_replay_frontier}} & \rev{\pageref{app:joint_replay_frontier}} & \rev{Joint-replay recipe and frontier} \\
 & \rev{\ref{app:sample_complexity_multiseed}} & \rev{\pageref{app:sample_complexity_multiseed}} & \rev{Sample complexity over training seeds} \\
 & \rev{\ref{app:joint_replay_safety}} & \rev{\pageref{app:joint_replay_safety}} & \rev{Trajectory-level safety and comfort} \\
 & \rev{\ref{app:distill}} & \rev{\pageref{app:distill}} & \rev{Distilled regional adapter} \\
\midrule
\rev{\S\ref{sec:variance_forensics}} & \rev{\ref{app:tta}} & \rev{\pageref{app:tta}} & \rev{Test-time adaptation} \\
 & \rev{\ref{app:hypernet}} & \rev{\pageref{app:hypernet}} & \rev{Hypernetwork-conditioned LayerNorm} \\
\midrule
\rev{\S\ref{sec:realvehicle}} & \rev{\ref{app:nuplan}} & \rev{\pageref{app:nuplan}} & \rev{nuPlan leave-one-city-out (open-loop)} \\
 & \rev{\ref{app:dpo_v3}} & \rev{\pageref{app:dpo_v3}} & \rev{Real-vehicle runs and the DPO adapter study} \\
 & \rev{\ref{app:vehicle}} & \rev{\pageref{app:vehicle}} & \rev{Test vehicle and its specifications} \\
\bottomrule
\end{tabular}
\end{table}
\FloatBarrier

\section{\rev{Test-Time Adaptation Details}}
\label{app:tta}

\subsection{\rev{TENT Recipes and Variance-Collapse Diagnostics}}
\label{sec:tta}

\rev{Table~\ref{tab:tta} reports TENT-style test-time adaptation under two recipes on the disjoint target-region evaluation pool of seed~$7$. The lightweight recipe, TENT-LN, updates only the affine parameters of the decoder LayerNorm layers. Its loss is the variance across $K{=}4$ sampled trajectories, which serves as the diffusion analogue of the prediction entropy used by TENT~\citep{Wang2020-io,Niu2022-zu}. TENT-LN runs $5$ SGD steps at $\mathrm{lr}{=}10^{-4}$, starting from a fresh model for each scene, and its trainable parameters make up $\sim\!0.01\%$ of the model. The full recipe, TENT-online, follows the defaults of Wang et al.\ (2021) where the architecture allows: Adam at $\mathrm{lr}{=}10^{-3}$, $K{=}32$ samples, model state carried over from scene to scene, and $10$ epochs over the stream of $24$ target-region scenes. We update LayerNorm rather than BatchNorm affine parameters because the Diffusion-Planner base planner has no BatchNorm layers; this is the standard substitution for transformer-based planners~\citep{Niu2022-zu}. A SAR variant backpropagates, in each epoch, only on the $50\%$ of scenes with the lowest variance.}

\emph{Collapse diagnostics.} \rev{The best TENT-online fleet mean, $4.11$\,m at epoch~$1$, comes from variance collapse rather than from learning the new region. Over ten epochs, the $K$-sample variance loss falls by four orders of magnitude ($40.6 \to 0.0095$), while the fleet mean rises to $5.09$\,m. For a diffusion trajectory planner under cross-region shift, any non-trivial TTA fleet mean should be reported together with (i) the within-scene variance $\sigma^2_K$ across $K$ samples and (ii) the pairwise $L_2$ distance between the mean trajectories of different scenes~\citep{Park2024-um,Yao2024-xr}. Healthy configurations have $\sigma^2_K \sim 10^{-11}$ and a cross-scene $L_2 \in [105, 150]$\,m. TENT started from the static default reaches $3.37$\,m, but its cross-scene $L_2$ collapses to $10.2$\,m. The hypernetwork-LN recipe (\S\ref{sec:hypernet_ln}) reduces $\sigma^2_K$ to $4{\times}10^{-13}$. Stacking TENT on the hypernetwork recipe gives a cross-scene $L_2$ of $1.2$\,m and $3.7$\,m of forward progress at $0.46$\,m/s.}

\begin{table}[h]
\centering
\footnotesize\setlength{\tabcolsep}{4pt}
\caption{\rev{Test-time adaptation baselines on the disjoint target-region evaluation pool of seed~$7$. TENT updates only the affine parameters of the decoder LayerNorm layers, using a trajectory-variance loss over $K$ samples. TENT-LN: $K{=}4$, $5$ SGD steps, fresh model for each scene. TENT-online: $K{=}32$, Adam with lr~$=10^{-3}$, $10$ epochs, model state carried over between scenes.}}
\label{tab:tta}
\resizebox{\ifdim\width>\linewidth\linewidth\else\width\fi}{!}{%
\begin{tabular}{lcccc}
\toprule
Method & Knowledge Hub & Ishiodai & Morikoro & Fleet mean \\
\midrule
\rev{Shared module only (TTA slice)}             & $14.26 \pm 4.13$ & $7.43 \pm 6.11$ & $37.40 \pm 29.68$ & $19.70$ \\
\rev{Shared module only (HyperLN slice)}         & ---              & ---             & ---               & $22.76$ \\
TENT-LN ($K{=}4$, $5$ steps)                     & $14.15 \pm 4.18$ & $7.47 \pm 6.14$ & $36.99 \pm 29.45$ & $19.54$ \\
\textbf{\rev{TENT-online ($K{=}32$, epoch~$1$)}} & ---              & ---             & ---               & $\mathbf{4.11}$ \\
\rev{TENT-online ($K{=}32$, epoch~$10$)}         & ---              & ---             & ---               & $5.09$ \\
\rev{TENT-online SAR ($K{=}32$, epoch~$4$)}      & ---              & ---             & ---               & $4.56$ \\
\midrule
\rev{Static default (ODD-Nagoya)}                & $4.51$ & $4.82$ & $31.88$ & $13.74$ \\
\rev{Joint replay (single checkpoint)}           & $\mathbf{2.24}$ & $\mathbf{2.77}$ & $\mathbf{7.78}$ & $\mathbf{4.26}$ \\
\bottomrule
\end{tabular}}
\end{table}

\FloatBarrier
\section{\rev{Hypernetwork-Conditioned LayerNorm}}
\label{app:hypernet}

\subsection{\rev{Recipe, Results and Diagnostic Probe}}
\label{sec:hypernet_ln}

\rev{This section describes a context-aware TTA recipe that trains a supervised mapping from the ECEF unit vector $\ell$ of the deployment location to offsets $(\Delta\gamma, \Delta\beta)$ of the $36{,}608$ LayerNorm affine parameters in the decoder DiT. The recipe reaches a low lateral offset, but the diagnostic probe of Table~\ref{tab:hypernet_diagnostics} shows that this offset is an artefact of nearly stationary trajectories.}

\paragraph{Recipe.} \rev{Let $\ell \in S^2$ be the ECEF unit vector of the deployment location (\S\ref{subsec:gps_mix}). The hypernetwork $h_\theta: S^2 \to \mathbb{R}^{36{,}608}$ is a three-layer MLP with hidden width $64$ and SiLU activations; it has $4.73$\,M parameters in total, and its output layer is zero-initialised. During training, each scene from a training ODD is paired with the ECEF unit vector $\ell_d$ of that ODD. At every forward pass we compute $(\Delta\gamma, \Delta\beta) = h_\theta(\ell_d)$ and apply these offsets to the LayerNorm modules through a forward hook that computes $F.\mathrm{layer\_norm}(x, w{+}\Delta\gamma, b{+}\Delta\beta, \varepsilon)$. We then run the diffusion sampler and minimise the mean-squared error to the recorded ego trajectory $\tau_{\mathrm{gt}}$. Training runs for $500$ optimiser steps at $\mathrm{lr}{=}5\!\cdot\!10^{-5}$ and $\mathrm{wd}{=}10^{-3}$ on a pool of $772$ pairs from the four training ODDs, and takes $2.7$ minutes of wall-clock time on one consumer GPU. At deployment, the trained hypernetwork is conditioned on the unit vector of the centroid of the target-region map, and its output is applied to the LayerNorm affine parameters.}

\paragraph{\rev{Results before the probe.}} \rev{Table~\ref{tab:hypernet} reports a fleet mean of $5.74$\,m for the best single seed ($6.27\pm0.59$ over three seeds). Reducing the hidden width ($64\to16$) raises the fleet mean to $11.0$\,m, which rules out a trivial lookup.}

\begin{table}[h]
\centering
\footnotesize\setlength{\tabcolsep}{4pt}
\caption{\rev{Lateral offset of hypernetwork-conditioned LayerNorm before the diagnostic probe of Table~\ref{tab:hypernet_diagnostics}.}}
\label{tab:hypernet}
\resizebox{\ifdim\width>\linewidth\linewidth\else\width\fi}{!}{%
\begin{tabular}{lcccc}
\toprule
Recipe & Knowledge Hub & Ishiodai & Morikoro & Fleet mean \\
\midrule
\rev{Shared module only}                          & $14.26$ & $7.72$ & $46.30$ & $22.76$ \\
\rev{Static default (ODD-Nagoya)}                 & $4.51$  & $4.82$ & $31.88$ & $13.74$ \\
\rev{Joint replay (single checkpoint)}            & $2.24$  & $2.77$ & $7.78$  & $4.26$ \\
\midrule
HyperLN, $h{=}64$, no jitter (best single seed)   & $11.19$ & $1.34$ & $4.68$ & $5.74$ \\
HyperLN, $h{=}64$, no jitter, $3$-seed mean       & $12.17 \pm 1.18$ & $1.39 \pm 0.20$ & $5.26 \pm 0.39$ & $6.27 \pm 0.59$ \\
HyperLN, $h{=}16$, jitter $0.1^\circ$ (capacity ablation) & $15.42$ & $1.95$ & $15.71$ & $11.02$ \\
\bottomrule
\end{tabular}}
\end{table}

\paragraph{Diagnostic probe.} \rev{The probe shows that the hypernetwork recipe produces nearly stationary trajectories. We measure forward progress $\|x(T) - x(0)\|$, mean speed $\|x(T) - x(0)\|/T$ at $T{=}8$\,s, and the cross-scene pairwise $L_2$ between trajectories, each centred at its own starting frame. Table~\ref{tab:hypernet_diagnostics} shows that the hypernetwork recipe produces $12$\,m of forward progress at $1.5$\,m/s (walking speed). Its low lateral offset is the trivial offset of a nearly stationary trajectory, so the apparent gains in Table~\ref{tab:hypernet} are artefacts.}

\begin{table}[h]
\centering
\footnotesize\setlength{\tabcolsep}{4pt}
\caption{\rev{Diagnostic probe of the hypernetwork-LN recipe (\S\ref{sec:hypernet_ln}) and of the TENT polish variants (\S\ref{sec:tta}). Forward progress is $\|x(T) - x(0)\|$ at $T{=}8$\,s; mean speed is progress$/T$. Cross-scene $L_2$ is the mean pairwise distance between the trajectories of different scenes, each centred at its own starting frame; healthy models give a large $L_2$.}}
\label{tab:hypernet_diagnostics}
\resizebox{\ifdim\width>\linewidth\linewidth\else\width\fi}{!}{%
\begin{tabular}{lcccc}
\toprule
Configuration                                  & lat (m) & progress (m, 8\,s) & speed (m/s) & cross-scene $L_2$ \\
\midrule
Frozen shared module only                              & $24.89$ & $170.0$ & $21.3$ & $150.4$ \\
Static default (ODD-Nagoya)             & $16.04$ & $166.7$ & $20.8$ & $105.0$ \\
vMF mixture $\tau{=}200$\,km               & $19.64$ & $172.2$ & $21.5$ & $124.2$ \\
\midrule
\textbf{Hypernet-LN (this work)}                & $5.27$  & $\mathbf{12.3}$ & $\mathbf{1.5}$ & $73.0$ \\
\textbf{Hypernet-LN $+$ TENT polish}            & $1.21$  & $\mathbf{3.7}$  & $\mathbf{0.46}$ & $\mathbf{1.2}$ \\
\textbf{Static default $+$ TENT polish}             & $3.40$  & $174.8$ & $21.9$ & $\mathbf{10.2}$ \\
\bottomrule
\end{tabular}}
\end{table}

\paragraph{\rev{Status of the hypernetwork recipe.}} \rev{We retract the hypernetwork-LN recipe as a constructive mechanism. The three constructive mechanisms reported in the main text are the vMF mixture (\S\ref{sec:gps_mix}), shared-adapter repair (\S\ref{sec:park_repair}), and joint replay (\S\ref{sec:joint_replay_frontier}). We include the hypernetwork-LN recipe as a documented failure mode, not as a deployable recipe, and it should not be cited as a constructive mechanism in later work.}

\FloatBarrier
\section{\rev{Derivation and Weights of the vMF Mixture}}
\label{app:pacbayes}

\paragraph{\rev{Bayesian interpretation of the mixing weights.}}
\rev{The mixing weights of the vMF mixture can be read as a Bayesian posterior over the training ODDs. Each weight is a softmax over the negative great-circle (Haversine) distance between the deployment location $\ell$ and the geographic centre $\mu_d$ of training ODD $d$:
\[
\alpha_d(\ell) \;=\; \frac{\exp\!\bigl(-\mathrm{d_{geo}}(\ell, \mu_d)/\tau\bigr)}{\sum_{j=1}^{K} \exp\!\bigl(-\mathrm{d_{geo}}(\ell, \mu_j)/\tau\bigr)},
\]
where the temperature $\tau$ is in kilometres. Let $D \in \{1, \ldots, K\}$ be the unobserved index of the training ODD, with a uniform prior, and let the deployment location have the likelihood $p(\ell \mid D = d) \propto \exp(-\mathrm{d_{geo}}(\ell, \mu_d)/\tau)$ on the sphere, which decays exponentially with distance from $\mu_d$. The normalising constant of this likelihood is invariant under rotations of the sphere, so it does not depend on $d$, and Bayes' rule gives $p(D = d \mid \ell) = \alpha_d(\ell)$ exactly. In the limit $\tau{\to}0$ the weights select the nearest ODD, which is the static default on our three maps; in the limit $\tau{\to}\infty$ they become uniform; intermediate values of $\tau$ lie between these two cases. We keep the name vMF mixture because each ODD contributes a kernel on the sphere centred at its location. A von Mises--Fisher likelihood $\exp(\kappa\,\mu_d^{\!\top}\ell)$, however, decays with the squared distance at small separations, so it defines a different weighting from the one we evaluate. Under quadratic loss, the Bayes-optimal trajectory given the scene and location $(x, \ell)$ is the posterior mean
\[
\hat{y}^{\star}(x, \ell) \;=\; \sum_{d=1}^{K} \alpha_d(\ell)\, f_d(x),
\]
where $f_d(x)$ is the expert of training ODD $d$. Our weight mixture $\widetilde{W}(\ell) = \sum_d \alpha_d(\ell)\,W_d$ averages the expert weights rather than their outputs, so it matches this rule only to first order in the differences between the expert weights. Table~\ref{tab:pacbayes_bound} lists the resulting weights.}

\begin{table}[h]
\centering\footnotesize\setlength{\tabcolsep}{4pt}
\caption{\rev{vMF mixture weights $\alpha$ and per-map lateral offset (m) across $\tau$.}}
\label{tab:pacbayes_bound}
\resizebox{\ifdim\width>\linewidth\linewidth\else\width\fi}{!}{%
\begin{tabular}{lcccc}
\toprule
$\tau$ & $\alpha$ \rev{(Hongo, Kashiwa, Shalun, Nagoya)} & Knowledge Hub & Ishiodai & Morikoro \\
\midrule
$\to 0$ (hard MAP $=$ ODD-Nagoya)     & $(0,\,0,\,0,\,1)$                & $4.51$ & $4.82$ & $31.9$ \\
$200$\,km (soft mix, headline)   & $(0.20,\,0.18,\,0.00,\,0.62)$    & $4.02$ & $4.63$ & $38.9$ \\
$\to \infty$ (uniform soup)      & $(0.25,\,0.25,\,0.25,\,0.25)$    & --- & --- & --- \\
\bottomrule
\end{tabular}}
\end{table}

\paragraph{\rev{Choice of distance metric.}} \rev{At our geographic scales, the mixing weights do not depend on the choice of distance metric. Table~\ref{tab:metric_robustness} reports the weights of the vMF mixture for Morikoro at $\tau{=}200$\,km under four distance metrics: Haversine, the ECEF chord, an equirectangular flat-Earth approximation and a local tangent plane (UTM). The four weight vectors agree to four decimal places (largest $L_1$ deviation $7{\times}10^{-5}$).}

\begin{table}[h]
\centering
\footnotesize\setlength{\tabcolsep}{4pt}
\caption{\rev{Mixing weights $\alpha$ of the vMF mixture for Morikoro at $\tau{=}200$\,km under four distance metrics.}}
\label{tab:metric_robustness}
\resizebox{\ifdim\width>\linewidth\linewidth\else\width\fi}{!}{%
\begin{tabular}{lcccc}
\toprule
Distance metric         & $\alpha_{\text{Hongo}}$ & $\alpha_{\text{Kashiwa}}$ & $\alpha_{\text{Shalun}}$ & $\alpha_{\text{Nagoya}}$ \\
\midrule
Haversine (great-circle)    & $0.1981$ & $0.1774$ & $0.0000$ & $0.6245$ \\
ECEF chord on $S^2$         & $0.1982$ & $0.1774$ & $0.0000$ & $0.6244$ \\
Equirectangular flat-Earth  & $0.1981$ & $0.1774$ & $0.0000$ & $0.6245$ \\
Local-tangent-plane (UTM)   & $0.1981$ & $0.1774$ & $0.0000$ & $0.6245$ \\
\bottomrule
\end{tabular}}
\end{table}

\rev{Table~\ref{tab:gps} reports the lateral offset of the vMF mixture, with the posterior weights derived above, at four temperatures.}

\begin{table}[h]
\centering
\footnotesize\setlength{\tabcolsep}{4pt}
\caption{\rev{Lateral offset (m) of the vMF mixture at four temperatures. As $\tau{\to}0$ the mixture reduces to the static default (all weight on ODD-Nagoya); as $\tau{\to}\infty$ it becomes the uniform average of the experts.}}
\label{tab:gps}
\resizebox{\ifdim\width>\linewidth\linewidth\else\width\fi}{!}{%
\begin{tabular}{lccc}
\toprule
$\tau$ (km) & Knowledge Hub & Ishiodai & Morikoro \\
\midrule
$1$ / $10$ / $50$ (one-hot $\to$ static default) & $4.51 \pm 0.79$ & $4.82 \pm 0.84$ & $\mathbf{31.9 \pm 32.1}$ \\
$200$ (soft mix)        & $\mathbf{4.02 \pm 0.11}$ & $4.63 \pm 0.68$ & $38.9 \pm 37.1$ \\
$200$ (top-2 mix)       & $4.13 \pm 0.73$ & $\mathbf{4.49 \pm 0.80}$ & $40.7 \pm 37.7$ \\
\bottomrule
\end{tabular}}
\end{table}

\FloatBarrier
\section{\rev{Weight-Averaging Baselines: Learned Coefficients, Model Soups and TIES-Merging}}
\label{app:weight_avg}

\paragraph{\rev{Learned mixing coefficients.}}\label{app:learned_coef}
\rev{Learning the mixing coefficients on target-region scenes does not beat the static default either. We replace the vMF weights with a learnable $4$-vector $\alpha$, $W_{\text{mix}} = W_{\text{base}} + \sum_d \mathrm{softmax}(\alpha)_d \cdot \theta_d$ with $\theta_d = W_d - W_{\text{base}}$, and train $\alpha$ for $200$ Adam steps at $\mathrm{lr}{=}10^{-2}$ on $20$ target-region scenes under the $10$-term reward (\S\ref{subsec:reward}), with the base planner frozen. The learned weights concentrate on one expert that is not the geographically nearest ($\alpha = (0.06, 0.12, \mathbf{0.70}, 0.12)$, entropy $1.34$\,bits). On the seed-$7$ evaluation pool, the resulting fleet mean is $22.82$\,m (Knowledge Hub $8.74$, Ishiodai $31.72$, Morikoro $28.00$), worse than the static default on the same pool ($17.86$\,m, Table~\ref{tab:greedy_soup}). No $\alpha$ that we tried within the convex hull of the four experts beats the static default; the larger gains in this paper come from shared-module repair (\S\ref{sec:park_repair}) and joint replay (\S\ref{sec:joint_replay_frontier}).}

\paragraph{\rev{Greedy soup.}}
\rev{Greedy soup~\citep{Wortsman2022-ra} over the four per-ODD experts ends with ODD-Nagoya alone (Table~\ref{tab:greedy_soup}). We use the seed-$7$ disjoint target-region pool for validation. Starting from the best single expert, ODD-Nagoya, the procedure rejects every candidate addition. The uniform soup over all four experts ($21.81$\,m) is worse than ODD-Nagoya alone ($17.86$\,m).}

\begin{table}[h]
\centering
\footnotesize\setlength{\tabcolsep}{4pt}
\caption{\rev{Greedy soup of the four per-ODD experts on the seed-$7$ disjoint target-region evaluation pool (CRC32 split, App.~\ref{app:hp}). The experts are ranked by fleet mean (lowest first), the soup is initialised with the best one, and each subsequent expert is added only if the uniformly averaged soup improves on the evaluation pool. In the trial rows, A, B, C and D denote ODD-Hongo, ODD-Kashiwa, ODD-Shalun and ODD-Nagoya. Reference rows give the uniform soup over all four experts and the vMF mixture at $\tau{=}200$\,km (GPS soft mix); the latter is copied from Table~\ref{tab:gps}, was measured on Pool~A, and is not comparable with the other rows.}}
\label{tab:greedy_soup}
\resizebox{\ifdim\width>\linewidth\linewidth\else\width\fi}{!}{%
\begin{tabular}{lcccc}
\toprule
Configuration                                       & Knowledge Hub  & Ishiodai  & Morikoro  & Fleet mean \\
\midrule
ODD-Hongo (ODD-Hongo) alone                                 & $5.99$ & $39.06$& $55.59$& $33.55$ \\
ODD-Kashiwa (ODD-Kashiwa) alone                               & $5.98$ & $31.52$& $22.90$& $20.13$ \\
ODD-Shalun (ODD-Shalun) alone                                & $13.6$ & $31.53$& $30.63$& $25.25$ \\
ODD-Nagoya (ODD-Nagoya) alone                               & $\mathbf{4.51}$ & $\mathbf{31.58}$& $\mathbf{17.48}$& $\mathbf{17.86}$ \\
\midrule
Soup \{D, B\} (trial)                                & $4.62$ & $31.80$& $19.93$& $18.78$  \\
Soup \{D, C\} (trial)                                & $4.79$ & $29.90$& $22.30$& $19.00$ \\
Soup \{D, A\} (trial)                                & $3.79$ & $34.12$& $31.09$& $23.00$ \\
Uniform soup over all four                          & $4.93$ & $32.91$& $27.60$& $21.81$ \\
\midrule
\textbf{Greedy soup final (single expert ODD-Nagoya)}     & $\mathbf{4.51}$ & $\mathbf{31.58}$ & $\mathbf{17.48}$ & $\mathbf{17.86}$ \\
GPS soft mix $\tau{=}200$\,km (reference, Table~\ref{tab:gps}) & $4.02$ & $4.63$ & $38.9$ & --- \\
\bottomrule
\end{tabular}}
\end{table}

\begin{figure}[h]
\centering
\includegraphics[width=0.97\linewidth]{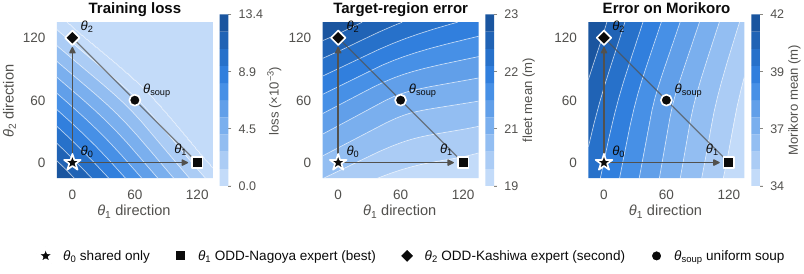}
\caption{\rev{Two-dimensional slice of the loss/error landscape spanned by two per-ODD experts, drawn in the style of Figure~2 of Wortsman et al.~\citep{Wortsman2022-ra}. The anchors are $\theta_0$ (shared module only; fleet mean $20.11$\,m, Morikoro $39.67$\,m), $\theta_1$ (ODD-Nagoya; $19.41$, $34.94$), $\theta_2$ (ODD-Kashiwa; $22.96$, $41.16$) and $\theta_{\mathrm{soup}}{=}\tfrac12(\theta_1+\theta_2)$ ($20.70$, $37.64$). Only these four points are evaluated; the surface is a thin-plate-spline interpolation through them.}}
\label{fig:landscape}
\end{figure}

\paragraph{\rev{Interpolation advantage of uniform soups.}}
\rev{Uniform soups usually beat the average of their endpoints but not the best endpoint. We adapt the interpolation-advantage metric of Wortsman et al.~\citep{Wortsman2022-ra} to errors, where lower is better:
\[
\Delta_{\mathrm{I}}(W_1, W_2) \;\triangleq\; \tfrac{1}{2}\bigl(\mathrm{err}(W_1) + \mathrm{err}(W_2)\bigr) \;-\; \mathrm{err}\!\bigl(\tfrac{1}{2}W_1 + \tfrac{1}{2}W_2\bigr).
\]
$\Delta_{\mathrm{I}}$ is positive if and only if the soup beats the arithmetic mean of its endpoints, which indicates a bonus from linear mode connectivity~\citep{Frankle2020-jj}. For the soup to also beat the best endpoint (the strict soup effect of Wortsman et al.), $\Delta_{\mathrm{I}}$ must exceed the inter-task gap, defined as the mean endpoint error minus the best endpoint error. Equivalently, the gap to the best endpoint, $\Delta_{\mathrm{gap}}$, must be negative. $\Delta_{\mathrm{I}}$ is positive in nearly every cell of Table~\ref{tab:interp_adv}, but in fleet mean the inter-task gap is $1.4$ to $5.4$ times larger. The soup therefore loses to the best single expert except in two cells (visualised in Figure~\ref{fig:landscape}).}

\begin{table}[h]
\centering
\footnotesize\setlength{\tabcolsep}{4pt}
\caption{\rev{Interpolation advantage $\Delta_{\mathrm{I}}$ (mean endpoint error minus soup error) for each two-expert and four-expert uniform soup on the seed-$7$ CRC32-disjoint target-region evaluation pool (App.~\ref{app:hp}). Positive values indicate that the soup beats the arithmetic mean of its endpoints, a bonus from linear mode connectivity. Table~\ref{tab:interp_adv_gap} reports $\Delta_{\mathrm{gap}}$ for the same rows.}}
\label{tab:interp_adv}
\begin{tabular}{l c c c c}
\toprule
Soup                       & Knowledge Hub & Ishiodai & Morikoro & Fleet \\
\midrule
$\{$ODD-Nagoya, ODD-Kashiwa$\}$        & $+0.63$ & $-0.26$ & $+0.26$ & $+0.21$ \\
$\{$ODD-Nagoya, ODD-Shalun$\}$        & $+4.27$ & $+1.66$ & $+1.76$ & $+2.56$ \\
$\{$ODD-Nagoya, ODD-Hongo$\}$        & $+1.46$ & $+1.20$ & $+5.45$ & $+2.70$ \\
Uniform $\{$D, B, C, A$\}$  & $+2.60$ & $+0.51$ & $+4.06$ & $+2.39$ \\
\bottomrule
\end{tabular}
\end{table}

\begin{table}[h]
\centering
\footnotesize\setlength{\tabcolsep}{4pt}
\caption{\rev{Gap to the best endpoint, $\Delta_{\mathrm{gap}}$ (soup error minus the error of the best single endpoint; positive values mean the soup is worse), for the same soups as Table~\ref{tab:interp_adv} on the seed-$7$ CRC32-disjoint target-region evaluation pool (App.~\ref{app:hp}). Bold cells have negative $\Delta_{\mathrm{gap}}$: the soup beats the best single expert, the strict soup effect of Wortsman et al.}}
\label{tab:interp_adv_gap}
\begin{tabular}{l c c c c}
\toprule
Soup                       & Knowledge Hub & Ishiodai & Morikoro & Fleet \\
\midrule
$\{$ODD-Nagoya, ODD-Kashiwa$\}$        & $+0.10$ & $+0.29$ & $+2.45$ & $+0.93$ \\
$\{$ODD-Nagoya, ODD-Shalun$\}$        & $+0.28$ & $\mathbf{-1.63}$ & $+4.82$ & $+1.14$ \\
$\{$ODD-Nagoya, ODD-Hongo$\}$        & $\mathbf{-0.72}$ & $+2.54$ & $+13.61$ & $+5.14$ \\
Uniform $\{$D, B, C, A$\}$  & $+0.41$ & $+1.39$ & $+10.11$ & $+3.95$ \\
\bottomrule
\end{tabular}
\end{table}

\paragraph{\rev{TIES-Merging.}}
\rev{Sign-aware merging does not close the inter-task gap either (Table~\ref{tab:ties_merging}). TIES-Merging~\citep{Yadav2023-fs} reduces interference between task vectors in three steps: within each layer, each $\Delta_i$ is trimmed to its top-$k$\% entries by magnitude; for each entry, the elected sign is that of the magnitude-weighted majority; and the merged $\Delta$ averages only the entries whose trimmed sign matches the elected sign. We use keep-densities of $20\%$, the standard setting of Yadav et al.\ (``canonical Yadav'' in the table), and $50\%$ as an ablation. Across all expert pairs and adapter layers, the mean absolute off-diagonal cosine similarity between task vectors is $|\!\cos|{=}0.065$. The task vectors are therefore nearly orthogonal, so TIES has few sign conflicts to resolve. TIES is statistically indistinguishable from the uniform soup ($22.4{-}22.5$\,m versus $21.8{-}22.3$\,m, with $\sim\!0.5$\,m of cuDNN re-evaluation noise) and loses to the best single expert ($17.86$\,m).}

\begin{table}[h]
\centering
\footnotesize\setlength{\tabcolsep}{4pt}
\caption{\rev{TIES-Merging~\citep{Yadav2023-fs} of the four per-ODD experts at two keep-densities, evaluated on the seed-$7$ CRC32-disjoint target-region evaluation pool (App.~\ref{app:hp}). Reference rows are taken from Table~\ref{tab:greedy_soup} and from the joint-replay frontier of \S\ref{sec:joint_replay_frontier}.}}
\label{tab:ties_merging}
\resizebox{\ifdim\width>\linewidth\linewidth\else\width\fi}{!}{%
\begin{tabular}{lcccc}
\toprule
Configuration                                                & Knowledge Hub  & Ishiodai  & Morikoro  & Fleet mean \\
\midrule
Best single (ODD-Nagoya, Table~\ref{tab:greedy_soup})              & $\mathbf{4.51}$ & $\mathbf{31.58}$ & $\mathbf{17.48}$ & $\mathbf{17.86}$ \\
Uniform soup all 4 (Table~\ref{tab:greedy_soup})              & $4.93$ & $32.91$ & $27.60$ & $21.81$ \\
\midrule
TIES-Merging, keep-density $20\%$ (canonical Yadav)           & $5.83$ & $35.56$ & $26.14$ & $22.51$ \\
TIES-Merging, keep-density $50\%$ (ablation)                  & $5.19$ & $35.05$ & $26.88$ & $22.37$ \\
Uniform soup all 4 (TIES re-eval, cuDNN noise)               & $6.07$ & $32.30$ & $28.51$ & $22.29$ \\
\midrule
Joint-replay from-scratch (the rank-allocation sweep) -- reference & $1.94$ & $6.00$ & $6.81$ & $\mathbf{4.92}$ \\
\bottomrule
\end{tabular}}
\end{table}

\FloatBarrier
\section{\rev{Shared-Adapter Repair: Per-Expert Results and Perturbation Ablation}}
\label{app:park_repair}

\begin{table}[h]
\centering
\footnotesize\setlength{\tabcolsep}{4pt}
\caption{Per-expert lateral offset (m) on the three held-out maps before and after fine-tuning the shared module on procedurally perturbed training-ODD scenes.}
\label{tab:park_repair}
\resizebox{\ifdim\width>\linewidth\linewidth\else\width\fi}{!}{%
\begin{tabular}{lccc}
\toprule
Configuration                                  & Knowledge Hub            & Ishiodai            & Morikoro \\
\midrule
Static default (ODD-Nagoya) (original shared)             & $4.51 \pm 0.79$  & $4.82 \pm 0.84$  & $31.88 \pm 32.05$ \\
\midrule
shared module only, original                          & $14.3 \pm 4.1$   & $7.7 \pm 4.6$    & $46.3 \pm 35.3$ \\
shared module only, repaired                          & $6.0 \pm 2.2$    & $6.1 \pm 1.6$    & $\mathbf{21.4 \pm 26.0}$ \\
\midrule
ODD-Hongo + repaired-shared                        & $24.5 \pm 13.2$  & $17.6 \pm 6.4$   & $50.3 \pm 43.2$ \\
ODD-Kashiwa + repaired-shared                        & $7.9 \pm 0.5$    & $7.8 \pm 0.5$    & $\mathbf{11.1 \pm 10.5}$ \\
ODD-Shalun + repaired-shared                        & $6.0 \pm 2.1$    & $5.8 \pm 1.5$    & $18.3 \pm 24.9$ \\
ODD-Nagoya + repaired-shared                        & $11.0 \pm 3.6$   & $9.5 \pm 2.6$    & $12.2 \pm 8.2$ \\
\bottomrule
\end{tabular}}
\end{table}

\paragraph{\rev{Ablation of the perturbation components.}}
\rev{The perturbation used for shared-module repair has three components: a per-lanelet SE(2) rotation drawn uniformly from $\pm 45^\circ$, a cubic Bezier curvature warp with amplitude $\pm 3$\,m on each lane, and a random boundary drop that removes one of the two lane boundaries with probability $0.5$. Table~\ref{tab:park_repair_ablation} disables each component in turn, repeats the $500$-step fine-tune of the shared module under the same $10$-term reward, and reports the resulting lateral offset, including per-seed values on Morikoro, the open-park map.}

\begin{table}[h]
\centering
\footnotesize\setlength{\tabcolsep}{4pt}
\caption{\rev{Ablation of the perturbation components used for shared-module repair, with three independent seeds per ablation row; the baseline row is a single seed. Each ablation row repeats the same $500$-step fine-tune of the shared module with one perturbation component disabled.}}
\label{tab:park_repair_ablation}
\resizebox{\ifdim\width>\linewidth\linewidth\else\width\fi}{!}{%
\begin{tabular}{lcccc}
\toprule
Configuration & Knowledge Hub (seeds) & Ishiodai (seeds) & Morikoro (seeds) & Morikoro per-seed \\
\midrule
Baseline: rot + bezier + drop & $\mathbf{5.99 \pm 2.23}$ & $\mathbf{6.05 \pm 1.63}$ & $\mathbf{21.4 \pm 26.0}$ & single seed \\
\midrule
$-$ SE(2) rotation            & $6.05 \pm 3.37$ & $7.12 \pm 3.61$ & $33.13 \pm 7.24$ & $[41.4, 28.0, 29.9]$ \\
$-$ Bezier curvature warp     & $7.01 \pm 2.99$ & $7.53 \pm 2.69$ & $26.28 \pm 4.70$ & $[31.5, 22.4, 24.9]$ \\
$-$ Boundary drop             & $6.95 \pm 2.50$ & $7.56 \pm 2.19$ & $28.12 \pm 13.96$ & $[44.1, 22.0, 18.3]$ \\
\bottomrule
\end{tabular}}
\end{table}

\rev{SE(2) rotation contributes the most; the Bezier warp and the boundary drop contribute less, and by similar amounts. An earlier single-seed run suggested that disabling the boundary drop costs nothing. We retract that claim: the seed-to-seed standard deviation of $13.96$\,m on Morikoro without the boundary drop shows that the finding was driven by noise.}

\FloatBarrier
\section{\rev{Distilled Regional Adapter}}
\label{app:distill}

\subsection{Regional Adapter via Distillation}
\label{sec:distill}

\rev{Table~\ref{tab:adapter} reports the distilled regional adapter under the protocol of \S\ref{subsec:eval}. The student is a rank-$4$ LoRA on the frozen base planner, and the teacher is a soft mixture of the experts at $\tau{=}200$ km. The student is trained on $56$ scenes from the seed-$42$ training pool and evaluated on the disjoint seed-$7$ lanelet pool.}

\begin{table}[h]
\centering
\footnotesize\setlength{\tabcolsep}{4pt}
\caption{\rev{Distilled regional adapter on the three held-out maps, with training and evaluation scenes drawn from disjoint lanelets (\S\ref{subsec:eval}). Each cell gives the mean and standard deviation over the seed-$7$ evaluation pool of road lanelets. The adapter is a single $78$K-parameter rank-$4$ LoRA on the base planner, with no per-ODD modules.}}
\label{tab:adapter}
\resizebox{\ifdim\width>\linewidth\linewidth\else\width\fi}{!}{%
\begin{tabular}{lcccc}
\toprule
Map & Max lateral (m) & Forward progress (m) & \rev{Static default (m)} & \rev{Change} \\
\midrule
Knowledge Hub & $18.63 \pm 6.19$  & $156.80 \pm 2.34$  & \rev{$4.51$}  & \rev{$+313\%$} \\
Ishiodai      & $19.32 \pm 30.53$ & $149.61 \pm 25.40$ & \rev{$31.58$} & \rev{$-39\%$} \\
Morikoro      & $37.83 \pm 20.69$ & $151.00 \pm 14.18$ & \rev{$17.48$} & \rev{$+116\%$} \\
\bottomrule
\end{tabular}}
\end{table}

\rev{The distilled student follows the coarse route geometry (forward progress of $138$+\,m) but beats the static default only on Ishiodai and is worse on Knowledge Hub and Morikoro (Table~\ref{tab:adapter}). The Lipschitz argument below explains this result at rank $4$.}

\paragraph{\rev{Lipschitz capacity of the student and its empirical limit.}}
\rev{The disjoint-evaluation result follows from the function class of the student. A rank-$r$ LoRA student on a frozen base planner has a Lipschitz constant bounded by the product of per-layer operator norms plus a rank-dependent perturbation term. Let $f_T: \mathcal{X} \to \mathbb{R}^d$ denote the teacher (the $\alpha$-blended merged-expert dictionary) and $f_S$ the rank-$r$ student. Each LoRA layer perturbs the base linear map $W_0 \in \mathbb{R}^{d_{out} \times d_{in}}$ by $B A$, with $A \in \mathbb{R}^{r \times d_{in}}$ and $B \in \mathbb{R}^{d_{out} \times r}$. With the standard initialisation and the weight-decay-bounded training that we use, $\|B A\|_2 \le \beta\sqrt{r}$ for a layer-wise constant $\beta$, so the global Lipschitz upper bound across $L$ MLP layers is}
\[
\mathrm{Lip}(f_S) \;\le\; \prod_{\ell=1}^{L} \bigl(\|W_0^{(\ell)}\|_2 + \beta_\ell\sqrt{r}\bigr).
\]
\rev{The per-layer perturbation rank of the teacher is $K\,r_e = 32$ (four experts $\times$ rank $8$). Its Lipschitz upper bound is the same product with $\sqrt{r}$ replaced by $\sqrt{32}$, which is roughly $\sqrt{8}$ larger at every layer. Structurally, the student is a low-pass version of the teacher.}

\rev{The rank-$128$ Skyline of \S\ref{sec:skyline} ($32\times$ larger) still does not match the per-expert baseline, so the bottleneck is the supervision signal (trajectory MSE to the teacher) rather than $r$. Replacing this signal with the joint-replay recipe of \S\ref{sec:joint_replay_frontier} (single checkpoint) gives $2.24 / 2.77 / 7.78$\,m on Knowledge Hub / Ishiodai / Morikoro with a budget of $20$ target-region scenes, on the same evaluation pool.}

\FloatBarrier
\section{\rev{Joint-Replay Recipe and Frontier}}
\label{app:joint_replay_frontier}\label{app:joint_replay_extra}\label{sec:perturbation}\label{tab:perturbation}

\subsection{The Joint-Replay Frontier}
\label{sec:joint_replay_frontier_detail}

\rev{The methods of \S\ref{sec:gps_mix}--\S\ref{sec:distill} keep the experts frozen. They cause no forgetting by construction, but their target-region accuracy is limited to the disjoint-evaluation results of Table~\ref{tab:adapter}. Here we remove the freeze and measure what a small, targeted relaxation gains.}

\paragraph{Recipe.}
\rev{Joint replay unfreezes the full adapter stack (parameter inventory in App.~\ref{app:hp}) and trains it to maximise the original ODD-conditioned reward of \S\ref{subsec:reward}. The training pool combines $200$ training-ODD scenes ($50$ from each of the four training ODDs) with a small target-region pool from the seed-$42$ disjoint training set, at a fixed ratio of $10$:$1$ in favour of training-ODD scenes. We use AdamW with $\mathrm{lr}{=}5\!\times\!10^{-5}$ and a gradient accumulation factor of $4$, and stop after $500$ optimiser steps ($\sim\!5$ minutes of wall-clock time on one consumer GPU). We call the resulting set of weights the \emph{joint-replay frontier}, because a single setting, the target-region pool size $n_{\mathrm{target-region}}$, moves it between the frozen adapter stack and full retraining.}

\paragraph{\rev{Trajectory metrics beyond lateral offset.}} \rev{Table~\ref{tab:joint_replay_safety} (App.~\ref{app:joint_replay_safety}) reports lateral offset, $95$th-percentile jerk magnitude, $95$th-percentile lateral acceleration, the fraction of trajectories that leave the $\pm 3$\,m corridor, and forward progress over $8$ seconds on Pool~A, the $24$-scene routing pool. These are open-loop proxies from a single rollout, not closed-loop safety measurements. The absolute jerk values ($700$--$1300$\,m/s$^3$) are large because we report raw third-order finite differences, without the Savitzky--Golay smoothing that the V$3$ comfort proxy of \S\ref{sec:appendix_dpo_v3} applies before its ISO~2631 threshold check. Only relative comparisons between configurations are meaningful.}

\paragraph{\rev{Data budget and shared-module initialisation.}}
\rev{The joint-replay frontier is a family with two settings, $(n_{\mathrm{target-region}}, \text{shared init})$. The pool size $n_{\mathrm{target-region}}$ trades data budget for held-out accuracy. The initialisation of the shared module, either inherited from the pretrained model or trained from scratch, trades forgetting on the training ODDs for target-region accuracy. The distilled regional adapter (\S\ref{sec:distill}) lies off this frontier, on the axis with no target data: it causes no forgetting by construction and reaches a fleet mean of $25.26$\,m. With the inherited shared module (inherit-shared), $n_{\mathrm{target-region}}{=}20$ gives $15.43$\,m with $\sum\!\Delta_{\text{route\_dev}}{=}-0.04$\,m. With the shared module trained from scratch, the same $n$ falls in the $\mathbf{5\!-\!7}$\,m range but costs a $+11\!-\!16$\,m regression on the training ODDs. These numbers come from the rank sweep and use the evaluation slice with $30$ training scenes per map, which is harder than the per-seed slices of \S\ref{sec:joint_replay_frontier}. They can be compared with each other but not with the numbers in the main text (see the footnote in App.~\ref{app:sample_complexity_multiseed}).}

\begin{figure}[h]
\centering
\includegraphics[width=0.88\textwidth]{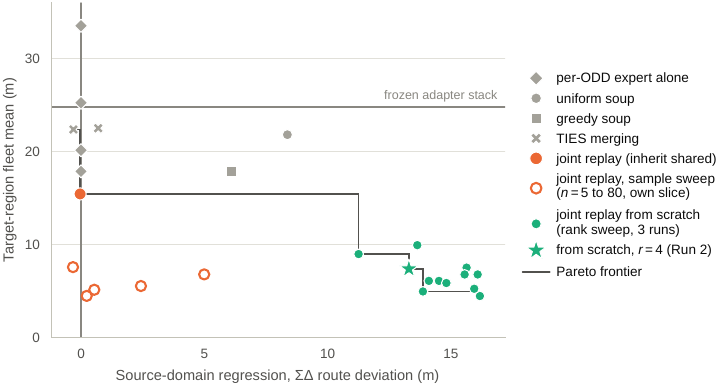}
\caption{\rev{Pareto view of target-region offset against forgetting for the weight-space baselines and the joint-replay configurations. Horizontal axis: regression on the training ODDs, measured as the increase in route deviation summed over the four training ODDs ($\Delta{=}0$ is the frozen adapter stack). Vertical axis: target-region fleet-mean lateral offset on the CRC32-disjoint seed-$7$ pool. Step line: empirical Pareto frontier. Star: the from-scratch $r{=}4$ checkpoint at $(+13.30, 7.35)$, which has not been deployed on the vehicle.}}
\label{fig:soup_pareto}
\end{figure}

\FloatBarrier
\section{\rev{Sample Complexity over Training Seeds}}
\label{app:sample_complexity_multiseed}

\subsection{Joint-Replay Sample Complexity}
\label{sec:sample_complexity}

\rev{Table~\ref{tab:sample_complexity} varies the target-region pool size $n_{\mathrm{target-region}} \in \{5,10,20,40,80\}$ for joint replay with the inherited shared module, with the training-ODD pool fixed at $200$ scenes and training fixed at $500$ optimiser steps. Target-region training scenes are drawn from the seed-$42$ disjoint training pool of $144$ candidate scenes. The evaluation set is the corresponding seed-$7$ pool of $24$ disjoint target-region scenes, built from the remaining lanelets.\footnote{Evaluation lanelets are drawn from the lanelets left after the training selection. The evaluation pool of this sweep ($n_{\mathrm{train\_per\_map}}{=}80$) is therefore a different subset of the seed-$7$ disjoint pool from the evaluation pool with $30$ training scenes per map, which is used by the joint-replay rank sweep of App.~\ref{app:joint_replay_extra} and the Stage~24-v2 inherit baseline. We document this dependence in App.~\ref{app:hp} and treat the two pools as separate evaluation slices of the same maps. Comparisons between this table and the rank-allocation sweep are therefore qualitative, not row-by-row quantitative.}}

\begin{table}[h]
\centering
\footnotesize\setlength{\tabcolsep}{4pt}
\caption{\rev{Sample complexity of joint replay with the inherited shared module (inherit-shared) on CRC32-disjoint seed-$7$ target-region evaluation slices ($24$ scenes per training seed, drawn from the lanelets left after that seed's training selection; see App.~\ref{app:hp}). Each row is a separate joint-replay training run with the given target-region pool size, a training-ODD to target-region ratio of $10$:$1$, and $500$ optimiser steps. Column 2: fleet mean for a single seed (seed-$42$). Column 3: fleet mean over seeds $\{42, 43, 44\}$, each scored on its own evaluation slice; the $n{=}80$ row has a single seed and is marked $\dagger$. The training-ODD columns sum over the four training ODDs.}}
\label{tab:sample_complexity}
\resizebox{\ifdim\width>\linewidth\linewidth\else\width\fi}{!}{%
\begin{tabular}{cccccc}
\toprule
$n_{\mathrm{target-region}}$ & fleet (seed-$42$, m) & fleet (3-seed mean$\pm$std, m) & Training-ODD $\sum\!\Delta_{\mathrm{route\_dev}}$ (m) & Training-ODD $\sum\!\Delta_{\mathrm{progress}}$ (m) & Train wall-clock \\
\midrule
Pre (shared module only; $n_{\mathrm{train}}{=}80$ eval slice) & $17.48$ & --- & --- & --- & --- \\
$\phantom{0}5$  & $6.76$ & $\phantom{0}9.79 \pm 5.14$ & $+5.00$ & $+25.86$ & $4.5$ min \\
$10$ & $7.54$ & $10.14 \pm 6.62$ & $\mathbf{-0.33}$ & $+3.29$ & $4.5$ min \\
$20$ & $5.51$ & $\phantom{0}6.17 \pm \mathbf{1.23}$ & $+2.43$ & $+21.37$ & $4.5$ min \\
$40$ & $5.10$ & $\phantom{0}5.67 \pm \mathbf{1.06}$ & $+0.54$ & $+4.69$ & $4.4$ min \\
$80^\dagger$ & $\mathbf{4.44}$ & --- ($\dagger$) & $+0.23$ & $-0.56$ & $4.3$ min \\
\bottomrule
\end{tabular}%
}
\end{table}

\paragraph{\rev{Seed variance at small $n$.}} \rev{At small $n$, the best result varies strongly across seeds. The three-seed std falls by $4.2\times$ from $n{=}5$ ($5.14$\,m) to $n{=}20$ ($1.23$\,m) (Figure~\ref{fig:sample_complexity}), and by a factor of five from $n{=}10$ ($6.62$\,m) to $n{=}20$. Over the same sweep, summed training-ODD forgetting stays within a cross-seed std of $0.94$--$1.23$\,m at every $n$. We therefore give $n_{\mathrm{target-region}}\!\geq\!20$ as the budget at which results are stable in practice. The single-seed minimum at $n{=}5$ (seed-$42$; $6.76$\,m, $-61\%$) is optimistic, since its three-seed mean is $9.79$\,m ($-44\%$). At $n{=}40$ the mean ($5.67\pm 1.06$\,m) is still above the single-seed $n{=}80$ reference ($4.44$\,m).}

\FloatBarrier
\section{\rev{Capacity Test with a High-Rank LoRA}}
\label{app:skyline}

\subsection{\rev{Capacity under Reward Supervision}}
\label{sec:skyline}

\rev{A LoRA with thirty-two times more parameters, trained by reward backpropagation, does not close the gap to the small adapter distilled with an MSE loss. This configuration, Skyline, attaches a rank-$128$ LoRA to every decoder MLP layer ($\approx 2.5$\,M trainable parameters, $\sim 32\times$ the distilled adapter), freezes the base planner, and overfits it to the held-out scenes by direct reward backpropagation for $2000$ steps. Table~\ref{tab:skyline} compares its lateral offset with that of the distilled adapter and the static default. The bottleneck is thus the supervision signal rather than capacity. Matching the teacher under an MSE loss is a low-curvature target that the Lipschitz-bounded student (\S\ref{sec:distill}) can track, whereas reward backpropagation admits reward maxima with large lateral offsets, which the high-capacity Skyline reaches first.}

\begin{table}[h]
\centering
\footnotesize\setlength{\tabcolsep}{4pt}
\caption{\rev{Skyline, an upper bound overfit to the held-out scenes, versus the deployable distilled adapter. Skyline: $2.5$\,M-parameter rank-$128$ LoRA trained for $2000$ optimiser steps; distilled adapter: $78$\,K-parameter rank-$4$ LoRA, $300$ optimiser steps, $1.5$\,min.}}
\label{tab:skyline}
\resizebox{\ifdim\width>\linewidth\linewidth\else\width\fi}{!}{%
\begin{tabular}{lccc}
\toprule
Configuration                                                     & Knowledge Hub & Ishiodai & Morikoro \\
\midrule
Static default (no adaptation)                              & $\mathbf{4.51}$ & $4.82$ & $31.88$ \\
\rev{Distilled regional adapter} ($78$\,K params, $1.5$\,min)             & $6.10$ & $\mathbf{4.05}$ & $\mathbf{18.88}$ \\
Skyline rank-$128$ LoRA ($2.5$\,M params, $\approx 15$\,min)       & $24.09 \pm 0.02$ & $24.28 \pm 0.09$ & $24.35 \pm 0.12$ \\
\bottomrule
\end{tabular}}
\end{table}

\FloatBarrier
\section{\rev{nuPlan Leave-One-City-Out Evaluation}}
\label{app:nuplan}

\subsection{\rev{Cross-City Evaluation on a Public Benchmark}}
\label{sec:nuplan}

\rev{We repeat the cross-city evaluation on the public nuPlan dataset with a leave-one-city-out protocol over four cities: Boston, Pittsburgh and Las Vegas (right-hand traffic, $\approx\!58$\,K scenes) and Singapore (sg-one-north, left-hand traffic, $\approx\!9$\,K scenes). Singapore is the only left-hand-traffic city in nuPlan.}

\paragraph{\rev{Planner variant and input/output changes.}} \rev{The nuPlan checkpoint is incompatible with our Autoware stack ($740$ missing keys under strict loading). We therefore use the nuPlan-compatible variant of the diffusion planner and reproduce our fine-tuning setup in it. A rank-$4$ shared LoRA is attached to every \texttt{nn.Linear} layer of the decoder outside multi-head attention ($89{,}472$ trainable parameters) and trained for $500$ steps with the official diffusion training loss (marginal-probability noise schedule and score MSE) on the three training cities (Boston, Pittsburgh and Las Vegas when Singapore is held out). No data from the held-out city is used. We make two input/output changes. The agent-future tensor of each scene is stored as $(x, y, \theta)$ and is converted to $(x, y, \cos\theta, \sin\theta)$ before the loss, and the $32$ neighbour-future tracks are truncated to \texttt{predicted\_neighbor\_num}${=}10$. The training loss decreases monotonically from $10.37$ to $3.38$ on the three-city pool.}

\paragraph{\rev{Three-seed $4{\times}4$ leave-one-city-out evaluation.}} \rev{For each held-out city, we fine-tune on the other three cities with three seeds $\{42,43,44\}$ and evaluate on a fixed pool of $50$ scenes per city. Table~\ref{tab:nuplan_4x4} reports the mean $\pm$ std over seeds; the std is below $1$\,m for every entry. Every held-out diagonal entry improves on the base planner in its column, and every entry for a training city also improves on the base, which shows that performance on the cities seen in training is preserved.}

\paragraph{\rev{Distance from the training cities.}} \rev{Singapore ($\approx\!1.3^\circ$\,N, left-hand traffic) lies $5{,}000$--$15{,}000$\,km from the three training cities. The held-out maps of the target region lie $11$--$17$\,km from ODD-Nagoya, but for Singapore a static default has no nearby ODD to fall back on. Even so, the Singapore-held-out row of Table~\ref{tab:nuplan_4x4} reduces the lateral offset on Singapore from $8.64$ to $5.60 \pm 0.78$\,m ($-35\%$) without any Singapore data. Because no data from the held-out city is used, this experiment does not test joint replay; it shows that the fine-tuned planner transfers to a distant city in open loop.}

\begin{table}[h]
\centering
\footnotesize\setlength{\tabcolsep}{4pt}
\caption{\rev{nuPlan $4{\times}4$ leave-one-city-out transfer matrix over three seeds. For each held-out city (row), a 3-city LoRA is trained on the other three cities with three independent seeds; each cell gives the mean $\pm$ std over seeds on one of the four cities (columns). Held-out diagonal entries are in bold. The base row (single seed) is the nuPlan BC checkpoint evaluated on the same fixed pool.}}
\label{tab:nuplan_4x4}
\resizebox{\ifdim\width>\linewidth\linewidth\else\width\fi}{!}{%
\begin{tabular}{lcccc}
\toprule
Configuration & Boston eval & Pittsburgh eval & Las Vegas eval & Singapore eval \\
\midrule
Base (single seed)                                  & $2.17$ & $4.51$ & $1.86$ & $8.64$ \\
\midrule
Boston held-out (3 seeds)                           & $\mathbf{1.63 \pm 0.16}$ & $3.15 \pm 0.24$ & $1.25 \pm 0.20$ & $5.10 \pm 0.30$ \\
Pittsburgh held-out (3 seeds)                       & $1.71 \pm 0.21$ & $\mathbf{3.40 \pm 0.37}$ & $1.37 \pm 0.27$ & $5.61 \pm 0.66$ \\
Las Vegas held-out (3 seeds)                        & $1.71 \pm 0.18$ & $3.27 \pm 0.21$ & $\mathbf{1.34 \pm 0.18}$ & $5.44 \pm 0.46$ \\
Singapore held-out (3 seeds, LHT)                   & $1.71 \pm 0.25$ & $3.49 \pm 0.44$ & $1.42 \pm 0.27$ & $\mathbf{5.60 \pm 0.78}$ \\
\bottomrule
\end{tabular}}
\end{table}

\paragraph{\rev{Single-seed $4{\times}4$ transfer matrix.}} \rev{To check whether the Singapore result is an artefact specific to that city, we also report the full $4{\times}4$ transfer matrix for a single seed (Table~\ref{tab:nuplan_matrix}). Held-out diagonal entries are within $80\%$--$99\%$ of the entries for cities seen during training. This is consistent with the fine-tuning transferring almost without loss across the three right-hand-traffic North American cities and partially to left-hand-traffic Singapore.}

\begin{table}[h]
\centering
\footnotesize\setlength{\tabcolsep}{4pt}
\caption{nuPlan $4{\times}4$ cross-city transfer matrix. Each row is a LoRA trained on three cities and evaluated on all four; the held-out city is shown in [brackets]. Maximum lateral offset (m), $50$ scenes per city.}
\label{tab:nuplan_matrix}
\resizebox{\ifdim\width>\linewidth\linewidth\else\width\fi}{!}{%
\begin{tabular}{lcccc}
\toprule
LoRA training set                       & Boston            & Pittsburgh         & Las Vegas          & Singapore \\
\midrule
None (base BC)                          & $2.14 \pm 6.68$ & $4.58 \pm 8.92$ & $1.80 \pm 3.95$ & $8.55 \pm 10.09$ \\
3-city, holdout Boston                  & $[1.71 \pm 4.55]$ & $\mathbf{3.03 \pm 5.94}$ & $1.36 \pm 2.50$ & $5.40 \pm 6.53$ \\
3-city, holdout Pittsburgh              & $1.71 \pm 4.24$ & $[3.50 \pm 6.42]$ & $\mathbf{1.34 \pm 2.21}$ & $5.41 \pm 6.51$ \\
3-city, holdout Las Vegas               & $\mathbf{1.62 \pm 4.28}$ & $3.36 \pm 6.48$ & $[1.25 \pm 2.31]$ & $\mathbf{5.30 \pm 6.46}$ \\
3-city, holdout Singapore               & $2.05 \pm 4.42$ & $3.69 \pm 6.37$ & $1.86 \pm 2.77$ & $[5.93 \pm 6.68]$ \\
\bottomrule
\end{tabular}}
\end{table}

\FloatBarrier
\section{Routing Feature Ablations}
\label{app:feature_ablation}

\rev{This section reports the three routing and feature experiments referenced in \S\ref{sec:routing_fails}. (i)~We extend the hand-crafted scene feature from $8$ to $38$ dimensions by adding histograms of traffic lights, line types and speed limits. The extended feature attains $1.000$ validation accuracy on the four-ODD training pool, but oracle routing accuracy drops from $0.375$ at $8$ dimensions to $0.250$ (chance) at $38$ dimensions, and the resulting fleet mean rises from $31.0$ to $36.9$\,m. (ii)~We measure how much information a feature carries about the per-scene oracle expert with a summed mutual-information statistic: we sum per-coordinate KSG estimates~\citep{Kraskov2004-ksg} over the feature dimensions. The sum is not the joint mutual information, which cannot exceed $\log 4 \approx 1.39$\,nats for a four-valued label, and it counts redundant information more than once. We therefore use it only as a test statistic against a permutation null. For the $256$-d planner-encoder embedding, the statistic drops from $60.3$ on the in-distribution training-ODD pool to $15.4$ on the held-out pool, against a permutation-null mean of $9.5$ ($p{=}0.088$). For the $3$-d GPS feature, it drops from $4.25$ to $0.00$ ($p{=}1.000$), because GPS coordinates vary little within the target region compared with the variation between training ODDs on which GPS was discriminative. (iii)~The reward-as-scorer Selector picks $\arg\max_e R(\tau_e)$ at inference time, using the planner's own training reward. It picks ODD-Shalun on $20/24$ held-out scenes, which gives a routing accuracy of $0.125$ (half that of random choice) and an offset of $29.8$\,m (mean over scenes; $26.6$\,m when averaged over maps as in Table~\ref{tab:routing}). The advantage of the static default comes from a constant marginal prior (always use the expert of the nearest training ODD), not from per-scene discrimination, regardless of whether the routing feature is based on perception or on geography.}
\FloatBarrier
\section{Hyperparameters and Training Recipes}
\label{app:hp}

\rev{Table~\ref{tab:hp} lists all training recipes; each runs on a single consumer GPU. All recipes use AdamW (weight decay $0.01$ unless noted otherwise), gradient-norm clipping at $1.0$ and gradient accumulation of $4$ (effective batch of $4$ scenes), without mixed precision. The DPM-Solver++ sampler runs $N{=}10$ second-order denoising steps with logSNR-spaced timesteps. It is deterministic and differentiable end to end, which enables reward back-propagation~\citep{Ren2024-hy, Zheng2024-gq}. No hyperparameter is tuned on the held-out maps; all values are inherited from the base Diffusion-Planner recipe or set by in-distribution validation.}

\begin{table}[h]
\centering
\footnotesize\setlength{\tabcolsep}{4pt}
\caption{\rev{Hyperparameters of the training recipes: optimisation budget. The budgets exclude the training of the underlying base planner, which is not part of this paper's contribution. The loss and training data of each recipe are given in Table~\ref{tab:hp_b}.}}
\label{tab:hp}
\resizebox{\ifdim\width>\linewidth\linewidth\else\width\fi}{!}{%
\begin{tabular}{lcccc}
\toprule
Recipe                                                   & Trainable params & LR  & Steps & Wall-clock \\
\midrule
Shared-module repair (\S\ref{sec:park_repair})            & $78{,}368$       & $5\!\times\!10^{-5}$ & $500$ & $3.9$ min \\
Rank-$8$ shared replication (App.~\ref{app:park_repair})  & $156{,}736$      & $5\!\times\!10^{-5}$ & $500$ & $4.0$ min \\
\rev{Regional-adapter distillation} (\S\ref{sec:distill})                & $78{,}368$       & $1\!\times\!10^{-4}$ & $300$ & $1.5$ min \\
Joint replay (\S\ref{sec:discussion})                     & $862{,}048$      & $5\!\times\!10^{-5}$ & $500$ & $5.0$ min \\
Joint-replay sample-complexity (\S\ref{sec:sample_complexity}) & $862{,}048$  & $5\!\times\!10^{-5}$ & $500$ & $4.4{-}5.7$ min each \\
Skyline rank-$128$ (\S\ref{sec:skyline})                  & $\sim\!2.5\!\times\!10^6$ & $5\!\times\!10^{-5}$ & $2000$ & $15$ min \\
VD-GRPO ($c{=}50$) (variance-decoupled GRPO study)        & $78{,}368$       & $5\!\times\!10^{-5}$ & $500$ & $13$ min \\
nuPlan $3$-city LoRA (\S\ref{sec:nuplan}, per LOO row)    & $89{,}472$       & $5\!\times\!10^{-5}$ & $500$ & $1$ min per row \\
nuPlan Singapore distillation                             & $89{,}472$       & $1\!\times\!10^{-4}$ & $200$ & $1$ min \\
\bottomrule
\end{tabular}%
}
\end{table}

\begin{table}[h]
\centering
\footnotesize\setlength{\tabcolsep}{4pt}
\caption{\rev{Hyperparameters of the training recipes: loss and training data (companion to Table~\ref{tab:hp}). In the training-data column, ``npz'' denotes scene files and ``prod'' denotes the training ODDs.}}
\label{tab:hp_b}
\begin{tabular}{p{0.18\linewidth}p{0.32\linewidth}p{0.41\linewidth}}
\toprule
Recipe                                                   & Loss & Train data \\
\midrule
Shared-module repair (\S\ref{sec:park_repair})            & ODD reward & $4{\times}50$ training-ODD npz, perturbed in-place \\
Rank-$8$ shared replication (App.~\ref{app:park_repair})  & ODD reward & same as above \\
\rev{Regional-adapter distillation} (\S\ref{sec:distill})                & MSE to teacher trajectory & $60$ seed-$42$ target-region train scenes \\
Joint replay (\S\ref{sec:discussion})                     & ODD reward & $200$ prod npz + $20$ target-region seed-$42$ \\
Joint-replay sample-complexity (\S\ref{sec:sample_complexity}) & ODD reward & $200$ prod + $\{5,10,20,40,80\}$ target-region \\
Skyline rank-$128$ (\S\ref{sec:skyline})                  & ODD reward & $24$ target-region eval scenes (overfit) \\
VD-GRPO ($c{=}50$) (variance-decoupled GRPO study)        & VD-GRPO advantage $\cdot$ noise-MSE log-pi & $200$ prod + perturbations \\
nuPlan $3$-city LoRA (\S\ref{sec:nuplan}, per LOO row)    & diffusion noise-MSE & $3000$ npz per non-held-out city \\
nuPlan Singapore distillation                             & MSE to teacher trajectory & $100$ Singapore teacher rollouts \\
\bottomrule
\end{tabular}
\end{table}

\paragraph{Reward function (per-ODD weights).}
\rev{The ODD-conditioned reward is a linear combination of $10$ terms that depend only on the trajectory, with separate weights for each ODD (\S\ref{subsec:reward}). The per-ODD weights are taken from the underlying Diffusion-Planner training run and are not tuned in this paper. As an example, the default weights are progress $2.0$, route deviation $2.0$, curvature $3.0$, comfort $0.5$, heading error $1.0$, stop bias $1.0$, lateral acceleration $1.0$, collision $4.0$, RSS obstacle $1.5$ and RSS route $0.5$. The progress term is clamped at $200$\,m so that the reward does not diverge on long-horizon predictions. At convergence, the reward typically lies in the $O(300{-}500)$ range.}

\paragraph{\rev{Disjoint training and evaluation split.}}
\rev{The training pool (seed $42$) and the evaluation pool (seed $7$) of road lanelets are produced by the deterministic helper function \texttt{disjoint\_scene\_split} in our release. For each held-out map, the helper (i) extracts all road lanelets, (ii) shuffles them with seed $42$ plus a per-map salt and takes the first $n_{\mathrm{train}}$ as training lanelets, and (iii) shuffles the remaining lanelets with seed $7$ and takes the first $n_{\mathrm{eval}}$ as evaluation lanelets. Knowledge Hub has only four road lanelets, so all four are reserved for evaluation and the map contributes no training scenes. The helper is unit-tested to guarantee zero overlap between the two index sets.}

\paragraph{Parameter-count breakdown of the base planner and adapter stack.}
\rev{Table~\ref{tab:param_breakdown} lists the parameters of the base planner, its DiT decoder and the adapter stack. LoRA wraps only the MLP \texttt{nn.Linear} layers of the DiT decoder, a subset of $4.05$\,M parameters. The remaining $\sim\!3.04$\,M DiT parameters (self-attention, cross-attention, LayerNorm and adaLN modulation) are frozen without LoRA modulation. They are not wrapped because the forward pass of PyTorch's \texttt{nn.MultiheadAttention} accesses \texttt{out\_proj.weight} through a fused-attention fast path, which fails on a wrapped \texttt{LoRALinear} child (see \S\ref{subsec:backbone}).}

\begin{table}[h]
\centering
\footnotesize\setlength{\tabcolsep}{4pt}
\caption{\rev{Parameter counts of the Diffusion-Planner base planner and of the adapter stack applied on top of it. Only the linear layers of the DiT MLPs are LoRA-wrapped; the attention and normalisation blocks stay frozen and are not modulated.}}
\label{tab:param_breakdown}
\resizebox{\ifdim\width>\linewidth\linewidth\else\width\fi}{!}{%
\begin{tabular}{lrr}
\toprule
Component                                                         &  Parameters & Role \\
\midrule
\multicolumn{3}{l}{\textbf{Base Diffusion-Planner (vanilla, no LoRA)}} \\
\quad encoder (frozen)                                            & $7{,}419{,}264$ & frozen \\
\quad decoder (frozen)                                            & $7{,}083{,}929$ & frozen \\
\quad\quad of which DiT total                                     & $7{,}082{,}564$ & frozen \\
\quad\quad\quad DiT.preproj (fc1 $+$ fc2)                          & $\phantom{0,0}297{,}728$ & LoRA target \\
\quad\quad\quad DiT.blocks ($\times 3$)                            & $5{,}922{,}816$ & mixed \\
\quad\quad\quad DiT.final\_layer (proj[1] $+$ proj[4])              & $\phantom{0,0}729{,}924$ & LoRA target \\
\quad\quad of which decoder wrapper/glue                          & $\phantom{0,000,}1{,}365$ & frozen \\
Full model total                                                  & $\mathbf{14{,}503{,}193}$ & frozen \\
\midrule
\multicolumn{3}{l}{\textbf{Per DiT block ($\times 3$, identical structure)}} \\
\quad self-attention                                              & $\phantom{0,0}263{,}168$ & frozen, no LoRA \\
\quad cross-attention                                             & $\phantom{0,0}263{,}168$ & frozen, no LoRA \\
\quad mlp1 (fc1 $+$ fc2)                                          & $\phantom{0,0}525{,}568$ & LoRA target \\
\quad mlp2 (fc1 $+$ fc2)                                          & $\phantom{0,0}525{,}568$ & LoRA target \\
\quad adaLN\_modulation                                            & $\phantom{0,0}394{,}752$ & frozen, no LoRA \\
\quad norm$\times 4$                                              & $\phantom{0,000,}2{,}048$ & frozen, no LoRA \\
\midrule
\multicolumn{3}{l}{\textbf{LoRA-wrapped frozen subset (the "modulated" base)}} \\
\quad sum over LoRA-target \texttt{nn.Linear} layers              & $\mathbf{4{,}046{,}404}$ & frozen base, modulated by LoRA \\
\midrule
\multicolumn{3}{l}{\textbf{adapter trainable modules (joint-replay configuration, \S\ref{sec:joint_replay_frontier})}} \\
\quad shared LoRA (rank $4$)                                       & $\phantom{0,00}78{,}368$ & trainable \\
\quad expert LoRA $\times 4$ (rank $8$ each)                       & $\phantom{0,0}626{,}944$ & trainable \\
\quad adapter trainable total                                       & $\mathbf{705{,}312}$ & trainable \\
\midrule
Full model with adapter                                            & $15{,}208{,}505$ & 95.4\% frozen, 4.6\% trainable \\
\bottomrule
\end{tabular}}
\end{table}

\FloatBarrier
\section{\rev{Details of the Routing Schemes}}
\label{app:routing_details}
\rev{This section gives the per-scheme details summarised in \S\ref{sec:routing_fails}. Unless stated otherwise, all results are on Pool~A.}

\rev{\textbf{Scene-conditioned MLP.} The MLP reaches $1.000$ validation accuracy on the four-ODD training pool. On the held-out maps, however, it routes nine of ten Ishiodai scenes and five of ten Morikoro scenes to ODD-Hongo rather than to the geographically nearest expert, ODD-Nagoya. This doubles the static default's offset on Ishiodai and on Morikoro.}

\rev{\textbf{Self-consistency.} Both variants, with and without magnitude normalisation, effectively select the shared module alone, because the consistency metric systematically favours small-magnitude outputs.}

\rev{\textbf{Denoising loss.} The pooled Kendall-$\tau$ correlation between the denoising loss and the realised offset of each expert is $0.172$, below the preregistered threshold of $0.2$.}

\rev{\textbf{Conformal corridor.} At $T{=}5$\,m, the corridor abstains on every scene. At $T{=}15$\,m, it reduces to the vMF mixture (\S\ref{sec:gps_mix}) and adds no triage signal of its own.}

\rev{\textbf{Reward-supervised gate.} The gate is trained directly on realised offset. Across $50$ seeds, it converges to a solution that places $0.967$ of its weight on ODD-Nagoya, and it matches the static default on Morikoro ($32.26 \pm 0.56$\,m).}

\rev{\textbf{Reward-as-scorer.} Selecting $\arg\max_e R(\tau_e)$ at inference time routes $20$ of $24$ scenes to ODD-Shalun. Its routing accuracy against the oracle is $0.125$, below the $0.25$ of random choice.}

\rev{\textbf{Oracle headroom and classifiers.} The oracle improves the mean offset over the $24$ scenes from $16.04$ to $15.15$\,m ($p{=}0.012$). This is a mean over scenes; Table~\ref{tab:routing} instead reports fleet means ($13.74$\,m for the same default). On a $104$-scene pool built from the same maps with the same protocol but new start lanelets, the gain is $1.34$\,m (bootstrap CI $[0.7, 2.2]$), and on Pool~B it is $1.1 \pm 0.5$\,m. For comparison, joint replay improves the mean over the $24$ scenes from $16.04$ to $6.95$\,m ($p{=}0.0005$; $19$ of $24$ scenes improve). Leave-one-out classifiers (logistic regression, $k$-NN and SVM) using the $8$ hand-crafted features or the $256$-d encoder embedding reach exactly the majority-class accuracy: $0.667$ on the $24$ scenes and $0.625$ on the $104$ scenes.}

\FloatBarrier
\section{Conformal Calibration Details}
\label{app:conformal}

\rev{Table~\ref{tab:conformal_appendix} reports the split-conformal calibration of each expert. For reproducibility, the last column also gives the in-sample residual quantile used in an earlier draft, which has no coverage interpretation.}

\begin{table}[h]
\centering
\footnotesize\setlength{\tabcolsep}{4pt}
\caption{\rev{Split-conformal calibration of each expert. For calibration, $40\%$ of each expert's own-ODD scenes are held out ($n{=}12$), and the split-conformal $q_e$ uses the finite-sample order statistic $\lceil (1-\alpha)(n{+}1) \rceil / n$ at $\alpha{=}0.1$. The last column reports the in-sample residual quantile, which has no coverage interpretation.}}
\label{tab:conformal_appendix}
\resizebox{\ifdim\width>\linewidth\linewidth\else\width\fi}{!}{%
\begin{tabular}{lcccccc}
\toprule
 & & & \multicolumn{2}{c}{\rev{Calibration residual (m)}} & \multicolumn{2}{c}{\rev{$q_e$ (m)}} \\
\cmidrule(lr){4-5}\cmidrule(lr){6-7}
Expert & $n_{\mathrm{train}}$ & $n_{\mathrm{cal}}$ & \rev{mean} & \rev{std} & \rev{split-conformal} & \rev{in-sample} \\
\midrule
ODD-Hongo    & $108$ & $12$ & $20.5$ & $20.0$ & $67.9$  & $83.5$ \\
ODD-Kashiwa  & $108$ & $12$ & $17.4$ & $11.0$ & $34.2$  & $53.9$ \\
ODD-Shalun   & $108$ & $12$ & $11.4$ &  $4.6$ & $17.1$  & $15.5$ \\
ODD-Nagoya   & $108$ & $12$ & $32.6$ & $20.4$ & $61.4$  & $33.4$ \\
\bottomrule
\end{tabular}%
}
\end{table}

\paragraph{Exchangeability permutation test.}
\rev{The calibration residuals on the training ODDs and the residuals on the target region fail a pooled exchangeability test. For each expert, we draw $B{=}10000$ permutations of the labels that mark each residual as a training-ODD calibration residual or a target-region residual, recompute the difference in means under each permutation, and report the two-sided $p$-value as the fraction of permutations whose absolute difference is at least the observed value. The pooled test, with $48$ training-ODD and $96$ target-region residuals, gives an observed difference of $+18.15$\,m and $p{=}0.0035$. Per-expert tests at $\alpha{=}0.05$ reject for ODD-Hongo ($p{=}0.045$) and ODD-Shalun ($p{=}0.0001$), but not for ODD-Kashiwa ($p{=}0.70$) or ODD-Nagoya ($p{=}0.094$). The residuals, the permutation-test results and a summary of the null distribution are in \texttt{results/routing\_pool/conformal\_exchangeability.json} of the released repository.}

\FloatBarrier
\section{\rev{Real-Vehicle Runs and a Separate DPO Adapter Study}}
\label{sec:appendix_dpo_v3}\label{app:dpo_v3}

\rev{This appendix reports a separate, deployment-oriented study on the same base planner and per-ODD experts. It is not part of the evaluation of joint replay or the other adaptation methods in the main text. The study fine-tunes a LoRA with a loss that combines Direct Preference Optimisation (DPO)~\citep{Rafailov2023-fs} with an L2 imitation anchor, and trains it inside an outer loop of Sharpness-Aware Minimisation (SAM)~\citep{Foret2020-al}. Every output trajectory is scored with a three-layer proxy for the JAMA preventability framework, whose layers cover hard safety, careful-and-competent driving, and comfort; the tables call this proxy V3. The resulting configuration is compared with a production DPO LoRA that has been validated on the real vehicle in ODD-Kashiwa, and with the reward-back-prop family of reward-trained adapters, whose regression in comfort motivated the study.}

\subsection{\rev{JAMA-Inspired Preventability Proxy}}
\label{subsec:v3_proxy}

\rev{The proxy is a per-trajectory scorer. It reports three layers separately and assigns each scenario to one of the buckets $\{$cold\_start, curve, straight, route\_following, object\_interaction$\}$. It is not a full implementation of JAMA Ver.~4.0~\S6, which requires a brake-delay simulator and a judge of Responder and Initiator roles; we build neither. A layer flags a trajectory if any of its criteria is triggered.}

\rev{The \texttt{strict\_safety} layer covers hard constraints with five criteria. These are a collision proxy ($0.5$\,m; this threshold is our own, since JAMA does not specify an absolute distance), a minimum time-to-collision ($2.0$\,s, taken from JAMA Ver.~4.0~\S6.3.1 Fig.~108, which cites UNR ECE/TRANS/WP.29/1091), departure from the route corridor ($1.75$\,m, derived from the $0.375$\,m wobble allowance in JAMA Fig.~102 plus a half-lane buffer), extreme deceleration ($9.81$\,m/s$^2$, the physical ceiling of $1$\,G; the JAMA cap for a careful and competent driver is much tighter at $0.774$\,G), and extreme lateral acceleration ($6.0$\,m/s$^2$, roughly $0.6$\,G). The \texttt{c\_and\_c\_driverlike} layer covers behaviour that a careful and competent driver would not produce, with six criteria: a burst of acceleration from a cold start, driver-like limits on lateral acceleration, deceleration and acceleration, inconsistency between heading and velocity, and a check for a stuck planner. The \texttt{comfort} layer covers degraded ride quality rather than safety, with three criteria: jerk above $5$\,m/s$^3$ (ISO 2631), lateral acceleration above $2$\,m/s$^2$, and a speed-change rate above $3$\,m/s$^2$.}

\rev{Kinematics are computed from raw positions after a Savitzky--Golay filter (window~$11$, polynomial order~$3$, \texttt{mode='interp'}). The filter removes the finite-difference noise floor of $\sim\!250$\,m/s$^3$ in raw $10$\,Hz position data. An earlier version of the proxy (v$1$) used \texttt{mode='nearest'} at the trajectory boundaries. This choice alone added a $32$\,m/s$^3$ artefact on linear motion and made the jerk criterion over-trigger on the recorded human baseline. We therefore switched to \texttt{mode='interp'}, which extrapolates the boundary samples with the fitted polynomial.}

\paragraph{\rev{Scene-level split of training and validation data.}} \rev{Training and validation data are split at the level of scenes rather than preference pairs. This fixes a problem found by an audit of an earlier iteration of this study (v$2$): the in-loop evaluator drew from all ODD-Kashiwa scenes, whether or not a scene also appeared in a training preference pair. With seed $2026$, the $3722$ ODD-Kashiwa scenes are split into $2977$ training and $745$ validation scenes, and an assertion checks that the two pools are disjoint. Tables~\ref{tab:v3_phase4} and \ref{tab:v3_prod} report results on the validation pool only, so no scene used for evaluation appears in a preference pair seen during training.}

\subsection{Hybrid DPO + L2 Imitation Loss and SAM Outer Loop}
\label{subsec:hybrid_dpo}

\rev{Preference pairs are built from ODD-Kashiwa training scenes. The \textbf{chosen} trajectory is the recorded ego trajectory (the \texttt{ego\_agent\_future} field of the scene file). The \textbf{rejected} trajectory is sampled from a reward-back-prop LoRA. We use three samplers, Reward-back-prop (joint-replay), Reward-back-prop, and Reward-LoRA soup (12-variant), and generate $\sim\!3000$ candidates with each. A pair is accepted if and only if the chosen trajectory passes the \texttt{c\_and\_c\_driverlike} layer of V3 and the rejected trajectory violates at least one criterion. This gives $9553$ accepted pairs out of $11166$ attempts ($85.6\%$) on $3722$ ODD-Kashiwa scenes. Pairs from the $745$ validation scenes are not used for training (see the scene-level split above).}

\rev{The hybrid loss interpolates between standard DPO and an imitation loss, the mean squared error to the chosen trajectory:}
\begin{equation}
\mathcal{L} \;=\; (1 - w_{\mathrm{imit}})\,\mathcal{L}_{\mathrm{DPO}}\;+\;w_{\mathrm{imit}}\,\mathcal{L}_{\mathrm{imit}}, \qquad \mathcal{L}_{\mathrm{imit}} \;=\; \|\hat{\tau}_\theta - \tau_{\mathrm{chosen}}\|_2^2 .
\label{eq:hybrid_dpo}
\end{equation}
\rev{Setting $w_{\mathrm{imit}}{=}0$ recovers vanilla DPO, and we show below that this configuration collapses. Setting $w_{\mathrm{imit}}{=}1$ recovers behaviour cloning with the L2 loss only. The DPO trust-region parameter $\beta$ controls how far the policy may drift from the reference distribution; a lower $\beta$ trusts the reference less and produces better-calibrated generations. The SAM outer loop perturbs the weights by $\rho\,\nabla_\theta \mathcal{L}/\|\nabla_\theta \mathcal{L}\|$ and re-evaluates the gradient at the perturbed point. This encourages convergence to a flat minimum, at $\sim\!1.9{\times}$ the wall-clock cost.}

\paragraph{\rev{Sweep results.}} \rev{We swept $29$ configurations in the second stage of the study (Phase~2). SAM is the most effective single regulariser. Any non-zero anchor weight in $[0.3,0.9]$ is stable, whereas pure DPO collapses (\S\ref{subsec:vanilla_dpo}). LoRA ranks~$4$ and $8$ work best; rank~$2$ collapses and ranks $\ge\!16$ degrade. For DPO, $\beta{=}0.01$~$>$~$0.1$~$>$~$1.0$. The source of the rejected trajectories affects the outcome substantially: Reward-back-prop alone gives a $3.3\%$ violation rate, against $40\%$ for Reward-back-prop (joint-replay). Across seeds, comfort varies by up to $30$\,pp.}

\paragraph{\rev{Selected configuration.}} \rev{The best of all $29$ trials, denoted DPO+L2+SAM, uses a shared module of rank $4$ and an expert of rank $8$ ($\alpha{=}32$), expert index $1$ (ODD-Kashiwa), the hybrid loss with $w_{\mathrm{imit}}{=}0.5$ and $\beta_{\mathrm{DPO}}{=}0.1$, AdamW (lr~$=3\!\cdot\!10^{-5}$, wd~$=10^{-4}$) with SAM ($\rho{=}0.05$), a constant learning-rate schedule, $10$ epochs with early stopping on the in-loop evaluation over the disjoint validation scenes, and $1500$ pairs.}

\subsection{\rev{Final Comparison (Phase 4) on 500 Validation Scenes}}
\label{subsec:v3_phase4}

\begin{table}[t]
\centering
\footnotesize\setlength{\tabcolsep}{4pt}
\caption{\rev{Final comparison (Phase~4) on $n{=}500$ ODD-Kashiwa validation scenes, disjoint from the training scenes, scored with the V3 proxy (\S\ref{subsec:v3_proxy}). Each entry is the percentage of the $500$ scenes in which at least one criterion of the layer is triggered (\texttt{any\_triggered}). \texttt{strict}: hard-constraint layer; \texttt{drv}: careful-and-competent driver layer; \texttt{cmf}: comfort layer.}}
\label{tab:v3_phase4}
\resizebox{\ifdim\width>\linewidth\linewidth\else\width\fi}{!}{%
\begin{tabular}{rlccc}
\toprule
Rank & Model & \texttt{strict}\,\% & \texttt{drv}\,\% & \texttt{cmf}\,\% \\
\midrule
1 & \textbf{DPO+L2+SAM} (this work, headline)             & $0.00$ & $100.0$ & $\mathbf{0.00}$ \\
1 & DPO+L2+SAM (block-restricted, LoRA on \texttt{blocks.0..2} only) & $0.00$ & $99.8$  & $\mathbf{0.00}$ \\
1 & greedy soup (degenerate; $\to$ \texttt{sam} alone) & $0.00$ & $100.0$ & $\mathbf{0.00}$ \\
4 & Base (no LoRA, Plan-R1 base planner)                        & $0.00$ & $97.2$  & $0.80$ \\
5 & DPO+L2 ($w_{\mathrm{imit}}{=}0.9$) & $0.00$ & $100.0$ & $3.80$ \\
6 & DPO+L2 ($\beta_{\mathrm{DPO}}{=}0.01$)       & $0.00$ & $100.0$ & $5.00$ \\
7 & DPO+L2+SAM (25 ep, final ckpt; $25$ ep $\times$ $3000$ pairs)      & $0.00$ & $99.8$  & $5.20$ \\
8 & Reward-LoRA (surgical zero, \texttt{decoder.dit.preproj.fc1})    & $0.00$ & $98.2$  & $\mathbf{7.40}$ \\
9 & DPO+L2+SAM (25 ep, best ckpt; early checkpoint)                    & $0.00$ & $98.8$  & $10.60$ \\
10 & DPO+L2 ($\alpha{=}64$)                       & $0.00$ & $100.0$ & $28.80$ \\
11 & Reward-back-prop (joint-replay)           & $0.00$ & $99.2$  & $79.80$ \\
12 & Reward-back-prop (ODD-Kashiwa expert)             & $0.00$ & $99.0$  & $81.40$ \\
13 & Reward-LoRA soup (12-variant Wortsman soup of reward-LoRAs) & $0.00$ & $99.2$  & $88.00$ \\
14 & DPO only ($w_{\mathrm{imit}}{=}0$; vanilla DPO, no anchor)        & $\mathbf{100.0}$ & $100.0$ & $\mathbf{100.0}$ \\
\bottomrule
\end{tabular}}
\end{table}

\begin{figure}[!htbp]
\centering
\includegraphics[width=\linewidth]{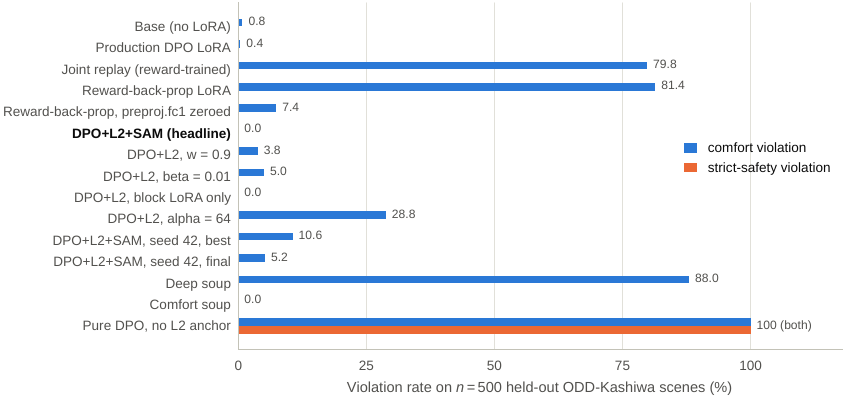}
\caption{\rev{Strict-safety and comfort-violation rates in the final comparison (Phase 4) on $n{=}500$ ODD-Kashiwa validation scenes, disjoint from the training scenes, under V3 (\S\ref{subsec:v3_proxy}). The comfort layer combines the \texttt{jerk}, \texttt{lat\_acc} and \texttt{speed\_change} criteria. Rows match the model rows of Table~\ref{tab:v3_phase4}; only the last model has strict-safety violations.}}
\label{fig:v3_phase4}
\end{figure}

\rev{The comfort column is the one that separates the models. The reward-back-prop family (Reward-back-prop, Reward-back-prop (joint-replay), Reward-LoRA soup (12-variant)) has comfort-violation rates of $79$--$88\%$, whereas DPO+L2+SAM, with the same LoRA architecture, reaches $0\%$. This isolates the fault to the fine-tuning method. As a further check, DPO+L2+SAM (block-restricted), with LoRA only on the block bodies and not on \texttt{preproj} or \texttt{final\_layer}, also reaches $0\%$, which is consistent with the per-layer ablation in \S\ref{subsec:per_layer}. The longer run of the third stage (Phase~3; $25$ epochs, $3000$ pairs) rises to $5.20\%$, which reflects overfitting rather than checkpoint noise.}

\subsection{\rev{Per-Layer Ablation and a Training-Free Fix}}
\label{subsec:per_layer}

\begin{figure}[!htbp]
\centering
\includegraphics[width=0.78\linewidth]{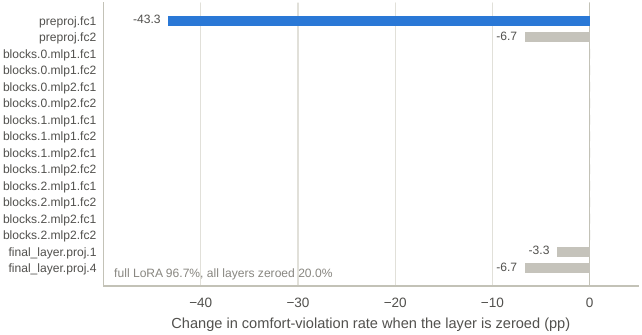}
\caption{\rev{Per-layer ablation of the Reward-back-prop LoRA. Each of the 16 LoRA-adapted DiT layers is zeroed in turn, and the comfort-violation rate is measured on $n{=}30$ held-out scenes.}}
\label{fig:per_layer}
\end{figure}

\rev{When a reward-back-prop LoRA is already deployed on a vehicle and full retraining is not affordable, a per-layer ablation offers a cheaper alternative. We zero each of the $16$ LoRA-adapted DiT layers of Reward-back-prop in turn and measure comfort on a held-out pool. A single layer stands out: \texttt{decoder.dit.preproj.fc1} accounts for a $-74$\,pp swing on its own, while every other layer contributes $7$\,pp or less. \texttt{preproj.fc1} is the first projection of the noisy input $x_t$ at the entry to the denoising chain. At this layer, the reward-back-prop gradient accumulates over all $80$ sampling steps and pulls the weights off the diffusion manifold. Zeroing only the LoRA update of this layer at inference, which takes three lines of code, lowers the comfort-violation rate of Reward-back-prop from $81.4\%$ to $7.4\%$ on the $n{=}500$ validation pool. This recovers about $90\%$ of the gap to the base model:}
\begin{verbatim}
for name, module in model.named_modules():
    if isinstance(module, adapterLinear) and 'preproj.fc1' in name:
        module.shared_scale = 0.0
        module.expert_scale = 0.0
\end{verbatim}
\rev{The resulting design rule is to keep the input and output projections of the DiT out of the LoRA target set and to place the rank budget on the block bodies.}

\subsection{\rev{Collapse of DPO Without the L2 Anchor}}
\label{subsec:vanilla_dpo}

\rev{Without the imitation anchor, DPO collapses. Setting $w_{\mathrm{imit}}{=}0$ reduces the hybrid loss to pure DPO. This configuration triggers $100\%$ on every V3 layer in every scenario bucket on $n{=}500$. Its median sampled jerk is $3627$\,m/s$^3$, against $0.8$ for the base planner, and its mean predicted speed is $46.6$\,m/s ($167$\,km/h), against $4$\,m/s for the base planner. The collapse is not an artefact of the jerk computation: it appears under the Savitzky--Golay-filtered jerk of V3 on $n{=}500$ and reproduces across the seeds we tried. Any non-zero anchor weight prevents the collapse. With $w_{\mathrm{imit}}{=}0.1$ alone, the comfort-violation rate drops to $53.3\%$, which is still poor but no longer catastrophic. With $w_{\mathrm{imit}}\ge 0.5$, it lies between $0\%$ and $13\%$, depending on the rest of the configuration.}

\subsection{\rev{Greedy Weight Soup over the Sweep Variants}}
\label{subsec:soup_degenerate}

\rev{A greedy soup over the $12$ variants from the sweep (Phase~$2$), with the strict-accept criterion of~\citep{Wortsman2022-ra}, reduces to DPO+L2+SAM alone. No other candidate strictly improves on its $0\%$ comfort-violation rate, so the greedy-soup row (marked degenerate) in Table~\ref{tab:v3_phase4} is identical to DPO+L2+SAM. The metric is saturated at $n{=}500$.}
\subsection{\rev{Test Vehicle}}
\label{app:vehicle}
\begin{figure}[h]
\centering
\begin{minipage}[c]{0.34\linewidth}
\centering
\includegraphics[width=\linewidth]{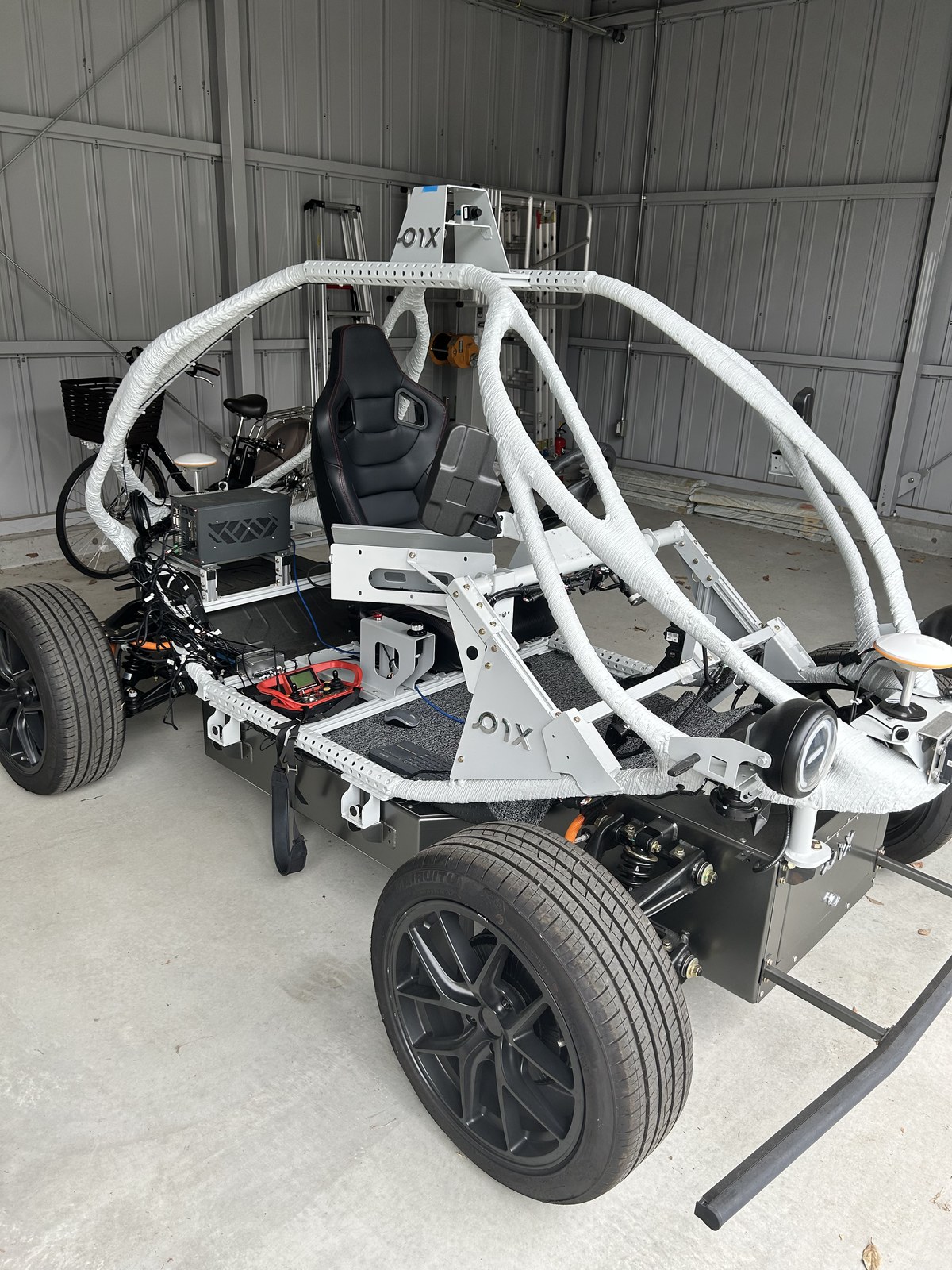}
\end{minipage}\hfill
\begin{minipage}[c]{0.62\linewidth}
\centering\footnotesize
\renewcommand{\arraystretch}{1.15}
\begin{tabular}{@{}ll@{}}
\toprule
\rev{Item} & \rev{Specification} \\
\midrule
\rev{Platform} & \rev{PIX Moving Pixkit~3.0, drive-by-wire} \\
\rev{Size (L$\times$W$\times$H)} & \rev{$2940 \times 1680 \times 1725$\,mm} \\
\rev{Wheelbase, track} & \rev{$1900$\,mm, $1455$\,mm} \\
\rev{Drive} & \rev{four $3$\,kW in-wheel motors} \\
\rev{Steering} & \rev{four-wheel; minimum turning radius $3.5$\,m} \\
\rev{Maximum speed} & \rev{$40$\,km/h (software limit)} \\
\rev{Vehicle interface} & \rev{CAN~2.0, $500$\,kbit/s} \\
\bottomrule
\end{tabular}
\end{minipage}
\caption{\rev{The test vehicle and its chassis specifications, as listed by the manufacturer.}}
\label{fig:vehicle}
\end{figure}

\rev{The real-vehicle runs in \S\ref{sec:realvehicle} use a PIX Moving Pixkit~3.0, an electric drive-by-wire chassis with an Autoware-compatible CAN interface (Figure~\ref{fig:vehicle}). On this vehicle, the base planner has been run with either a reward-trained shared-plus-expert adapter or the production DPO LoRA. The joint-replay weights have not been run on the vehicle.}

\subsection{\rev{Scoring Real-Vehicle Recordings with V3}}
\label{subsec:v3_realbag}

\begin{table}[t]
\centering
\footnotesize\setlength{\tabcolsep}{4pt}
\caption{\rev{V3 scores of ego trajectories recorded on the real vehicle. \texttt{rep1} baseline: a human driver drives a loop of the ODD-Kashiwa course. \texttt{rep1} (ODD-Kashiwa adapter): the planner, with the reward-trained shared-plus-expert adapter and the ODD-Kashiwa expert, drives the same course autonomously on the same day. \texttt{rep2} (ODD-Kashiwa adapter): a second autonomous run on the same day, in which the vehicle crept $72$\,m and then stopped without recovering.}}
\label{tab:v3_realbag}
\resizebox{\ifdim\width>\linewidth\linewidth\else\width\fi}{!}{%
\begin{tabular}{lrcccl}
\toprule
Bag & $n$ & \texttt{strict}\,\% & \texttt{drv}\,\% & \texttt{cmf}\,\% & Note \\
\midrule
\rev{\texttt{rep1} baseline}             & $27154$ & $0.1$ & $2.1$ & $\mathbf{37.7}$ & \rev{human driver} \\
\rev{\texttt{rep1} (ODD-Kashiwa adapter)} & $14275$ & $0.8$ & $5.2$ & $\mathbf{47.7}$ & \rev{adapter, autonomous} \\
\rev{\texttt{rep2} (ODD-Kashiwa adapter)} & $15755$ & $0.0$ & $0.0$ & $\mathbf{0.0}$  & \rev{crept $72$\,m, then stopped} \\
\bottomrule
\end{tabular}}
\end{table}

\paragraph{\rev{Comfort in simulation and on the vehicle.}} \rev{On the same course and day, the human driver has a comfort-violation rate of $37.7\%$ and the adapter-controlled run $47.7\%$. The deployed adapter is therefore $10$\,pp worse than the human driver it replaces. This matches the comfort regression of reward-back-prop training in simulation and motivated the switch to DPO+L2+SAM. In simulation, the planner's trajectories score below the human baseline (DPO+L2+SAM $0\%$, Base (no LoRA) $0.8\%$), because the output of the diffusion planner is smoother than raw human positions recorded at $10$\,Hz.}

\paragraph{\rev{Stalled second run (\texttt{rep2}).}} \rev{Row~3 reads $0.0\%$ because the vehicle never exceeded $0.23$\,m/s ($15755$ frames in the \texttt{cold\_start} bucket). We therefore add a progress check over the whole recording ($\bar{v}<0.5$\,m/s, $\ge 95\%$ stationary frames, $n\ge 300$). This check flags such a run as a deployment failure and complements V3. With the same adapter, on the same course and day, \texttt{rep1} was reasonable ($47.7\%$, $107$\,m driven), whereas in \texttt{rep2} the vehicle crept $72$\,m and then stopped.}

\subsection{\rev{Comparison with the Production DPO LoRA}}
\label{subsec:v3_prod}

\begin{table}[t]
\centering
\footnotesize\setlength{\tabcolsep}{4pt}
\caption{\rev{Comparison with the production DPO LoRA. DPO+L2+SAM (this work) is trained on explicit (chosen, rejected) preference pairs that are assembled automatically from the outputs of reward-back-prop samplers, without human labels. Production DPO (human labels) is the $3$-month production DPO LoRA, trained on human-labelled preference pairs and validated on the real vehicle in $2026$-$04$.}}
\label{tab:v3_prod}
\resizebox{\ifdim\width>\linewidth\linewidth\else\width\fi}{!}{%
\begin{tabular}{lllll}
\toprule
Model & Origin & \texttt{strict}\,\% & \texttt{cmf}\,\% & \rev{Real vehicle} \\
\midrule
\textbf{DPO+L2+SAM} (proposed)          & this work                  & $0.00$ & $\mathbf{0.00}$ & \rev{simulation only} \\
DPO+L2+SAM (block-restricted)           & this work                  & $0.00$ & $0.00$          & \rev{simulation only} \\
Production DPO (human labels)           & \rev{production, $3$ months} & $0.00$ & $\mathbf{0.40}$ & \rev{validated (April)} \\
Base (no LoRA)                          & Plan-R1 base planner       & $0.00$ & $0.80$          & \rev{simulation only} \\
Reward-back-prop                        & reward-back-prop           & $0.00$ & $81.40$         & \rev{failed (\texttt{rep2})} \\
Reward-LoRA (surgical zero, \S\ref{subsec:per_layer}) & this work, training-free & $0.00$ & $\mathbf{7.40}$ & \rev{not tested} \\
\bottomrule
\end{tabular}}
\end{table}

\rev{The proposed configuration matches the production DPO LoRA on hard safety ($0\%$) and is slightly better on comfort ($0.00\%$ vs.\ $0.40\%$), without human preference labels, because its preference pairs are assembled automatically. Deployment requires a LoRA delta of about $78$\,K parameters. When a deployed adapter cannot be retrained, the training-free surgical fix \texttt{v5\_surgical} (\S\ref{subsec:per_layer}) can be used instead.}

\rev{This comparison holds in simulation only. Two reference points from the real vehicle, the production DPO LoRA ($0.40\%$ in simulation) and the adapter-controlled \texttt{rep1} run ($47.7\%$ vs.\ $37.7\%$ for the human driver), place the comfort gap between simulation and vehicle on ODD-Kashiwa within $0$--$10$\,pp.}

\subsection{\rev{Adversarial Scenarios from AWSIM-Script}}
\label{subsec:awsim_adversarial}

\rev{In the natural ODD-Kashiwa data, $400/400$ sampled \texttt{neighbor\_agents\_past} tensors contain no active neighbours. We therefore synthesise dangerous interactions from the AWSIM-Script library~\citep{Tran2025-db}, which provides $42$ cut-in, $8$ cut-out and $4$ deceleration scripts. Each script is parameterised by $(V_e, V_o, V_y, dx_0)$: the ego speed, the NPC speed, the lateral speed of the NPC and the initial longitudinal gap. For each ODD-Kashiwa scene, we insert one synthetic NPC into slot~$0$ of \texttt{neighbor\_agents\_past}, with a three-second history at constant speed $V_o$ and a lane offset of $\pm 3.5$\,m for cut-in and cut-out, or $y{=}0$ for deceleration. We set the ego speed at $t{=}0$ to $V_e$ and score the planned trajectory against the analytically projected NPC future. The evaluation uses $n{=}200$ scenes per scenario and $4$ models, for $43{,}200$ inferences in total.}

\begin{table}[t]
\centering
\footnotesize\setlength{\tabcolsep}{4pt}
\caption{\rev{Collision rates in the AWSIM-Script adversarial scenarios, aggregated over the full $54$-script library, with $n{=}200$ ODD-Kashiwa scenes per scenario. Cut-in aggregates $42$ scripts spanning $V_e\in[20,50]\text{km/h}$, $V_o\in[10,40]\text{km/h}$, $V_y\in[0.6,1.6]\text{m/s}$ and $dx_0\in[5,17]\text{m}$; cut-out aggregates $8$ scripts; deceleration aggregates $4$, in which the NPC brakes at $1$\,G from $V_o\in\{20,30,40,50\}\text{km/h}$. Entries are the mean collision~\% across the scripts of each family. Table~\ref{tab:awsim_adv_canonical} gives the results for the five canonical scripts.}}
\label{tab:awsim_adv}
\begin{tabular}{lrrr}
\toprule
Model & \texttt{cutin}\,(42) & \texttt{cutout}\,(8) & \texttt{decel}\,(4) \\
\midrule
Base (no LoRA)                       & $\mathbf{17.0}$ & $\mathbf{0.2}$ & $\mathbf{48.0}$ \\
Reward-back-prop (reward-LoRA)             & $39.8$          & $4.5$          & $67.0$          \\
Reward-LoRA (surgical zero) (\S\ref{subsec:per_layer}) & $27.4$    & $0.7$          & $59.1$          \\
DPO+L2+SAM (ours)                 & $18.5$          & $1.0$          & $53.5$          \\
\bottomrule
\end{tabular}
\end{table}

\begin{table}[t]
\centering
\footnotesize\setlength{\tabcolsep}{4pt}
\caption{\rev{Per-script results on the five canonical scripts from the original AW-Runtime-Verification publication, with $n{=}200$ ODD-Kashiwa scenes per script, for comparison with the earlier pilot with $n{=}50$. The models are those of Table~\ref{tab:awsim_adv}.}}
\label{tab:awsim_adv_canonical}
\begin{tabular}{lrrrrr}
\toprule
Model & \texttt{ci30-10-6} & \texttt{ci40-20-3} & \texttt{ci20-10-1} & \texttt{co50-1} & \texttt{decel20} \\
\midrule
Base (no LoRA)                       & $\mathbf{35.0}$ & $\mathbf{17.0}$ & $\mathbf{41.0}$ & $\mathbf{0.0}$ & $\mathbf{73.5}$ \\
Reward-back-prop (reward-LoRA)             & $63.0$          & $39.0$          & $65.5$          & $0.0$ & $81.5$ \\
Reward-LoRA (surgical zero) (\S\ref{subsec:per_layer}) & $50.5$    & $30.0$          & $54.0$          & $0.0$ & $78.5$ \\
DPO+L2+SAM (ours)                 & $40.0$          & $17.5$          & $42.5$          & $0.0$ & $82.0$ \\
\bottomrule
\end{tabular}
\end{table}

\begin{figure}[!htbp]
\centering
\includegraphics[width=0.74\linewidth]{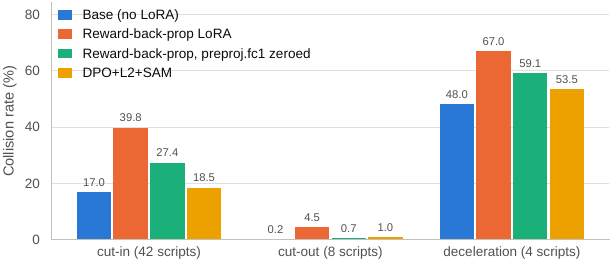}
\caption{\rev{Mean collision~\% over the full AWSIM-Script library of $54$ scenarios ($n{=}200$ ODD-Kashiwa scenes per scenario, $43{,}200$ inferences), aggregated by family (cut-in: $42$ variants; cut-out: $8$; deceleration: $4$).}}
\label{fig:awsim_adv}
\end{figure}

\begin{figure}[!htbp]
\centering
\includegraphics[width=0.91\linewidth]{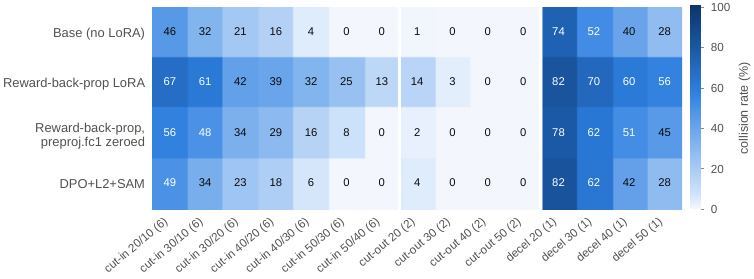}
\caption{\rev{Collision~\% for each scenario of the AWSIM-Script library ($54$ scripts, $n{=}200$ scenes per scenario). Rows are models; columns are the $54$ scripts, grouped by family and ego speed (number of scripts in parentheses). Darker cells have higher collision rates.}}
\label{fig:awsim_heatmap}
\end{figure}

\rev{In the adversarial scenarios, the strict-safety layer (\texttt{strict\_safety}) separates the models. On cut-in, aggregated over $42$ scripts, Reward-back-prop has a collision rate of $39.8\%$, compared with $17.0\%$ for the base planner ($+22.8$\,pp). Reward-LoRA (surgical zero) reduces this to $27.4\%$ without retraining, and DPO+L2+SAM stays close to the base planner at $18.5\%$. On deceleration, the collision rate is uniformly high for all models ($48$--$67\%$). This reflects a lack of training data, since the ODD-Kashiwa pool contains no emergency-braking examples; \S\ref{subsec:awsim_adv_train} addresses this.}

\begin{figure}[!t]
\centering
\includegraphics[width=0.93\linewidth]{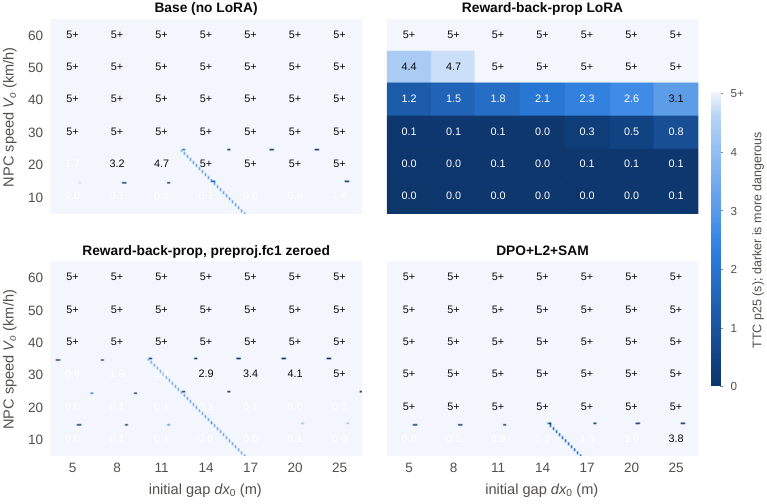}
\caption{\rev{Heatmap of $TTC_{p25}$ in the cut-in scenarios on a $7$ ($dx_0$, m) $\times$ $6$ ($V_o$, km/h) grid, with $V_y{=}1.6$\,m/s, $V_e{=}1.5\,V_o$ and $n{=}15$ ODD-Kashiwa scenes per cell. Cell labels report TTC p25 in seconds (5+ means at least $5$\,s); darker cells are more dangerous.}}
\label{fig:awsim_ttc}
\end{figure}

\paragraph{\rev{Time-to-collision heatmap.}} \rev{Figure~\ref{fig:awsim_ttc} reports $TTC_{p25}$ over a $7{\times}6$ $(dx_0, V_o)$ grid ($dx_0\in[5,25]$\,m, $V_o\in[10,60]$\,km/h, $V_e{=}1.5 V_o$, $V_y{=}1.6$\,m/s, $n{=}15$ scenes per cell). The heatmap is markedly darker, that is, more dangerous, for Reward-back-prop than for DPO+L2+SAM, with Reward-LoRA (surgical zero) in between.}

\paragraph{\rev{Limitations.}} \rev{The evaluation modifies one slot of the planner input and projects the NPC future analytically. It skips the perception and behaviour stages of AWSIM-Labs and the AW-Checker. Collisions are detected with a proxy based on a $0.5$\,m circular footprint. The values in Table~\ref{tab:awsim_adv} should therefore be read as relative comparisons, not as absolute rates.}

\subsection{\rev{Training with AWSIM-Script Adversarial Pairs}}
\label{subsec:awsim_adv_train}

\rev{The deceleration failures in Table~\ref{tab:awsim_adv} (every model at $48$--$67\%$) reflect a lack of training data. ODD-Kashiwa contains no scenarios in which a leading vehicle stops, so no fine-tuned model has seen reactive braking. We therefore generate adversarial preference pairs from the AWSIM-Script library and add them to the natural-data pool.}

\rev{The pairs use the same NPC insertion as \S\ref{subsec:awsim_adversarial}. For each AWSIM-Script script, we take $12$ ODD-Kashiwa training scenes, insert the NPC, and pair each modified scene with two trajectories. The rejected trajectory is sampled from DPO+L2+SAM itself. The pair is kept only if this sampled trajectory collides or has a TTC below $2$\,s, which removes scripts on which DPO+L2+SAM is already safe and keeps only the hard cases. The chosen trajectory is constructed analytically. For cut-in, the ego brakes from $V_e$ to $V_o$ at a comfort-capped $4$\,m/s$^2$. For deceleration, it brakes at the rate required to stop with a $5$\,m safety gap behind the NPC's stopping position. For cut-out, it continues at $V_e$, because the NPC moves out of the ego's path. The filter accepts $162$ adversarial pairs out of $648$ attempts across the $54$ scripts. Together with $1500$ natural ODD-Kashiwa pairs, they form a training pool of $1662$ pairs ($\sim\!10\%$ adversarial).}

\rev{Training uses the headline configuration of \S\ref{subsec:hybrid_dpo} (lr~$=3{\cdot}10^{-5}$, $w_{\mathrm{imit}}{=}0.5$, weight decay $10^{-4}$, SAM, constant schedule, $8$ epochs). It is initialised from the DPO+L2+SAM checkpoint rather than from the base planner, so that training starts from the existing $0\%$ comfort-violation rate on natural data.}

\paragraph{\rev{Comfort on natural data.}} \rev{Comfort on natural data is preserved at the end of training. During the eight training epochs, the in-loop comfort-violation rate on $n{=}30$ disjoint validation scenes is $23.3\%$ at epoch~$2$, immediately after the adversarial pool enters the gradient. It falls to $0.0\%$ at epoch~$6$ and stays at $0.0\%$ through epoch~$8$, with a median jerk of $1.3$\,m/s$^3$. The final checkpoint, DPO+L2+SAM+Adv, therefore matches DPO+L2+SAM on natural-data comfort (both $0\%$, Table~\ref{tab:v3_phase4}), while its LoRA delta also covers the adversarial pool.}

\begin{table}[h]
\centering
\footnotesize\setlength{\tabcolsep}{4pt}
\caption{\rev{Training with adversarial pairs from the AWSIM-Script library, for two sizes of the adversarial pool. Evaluation uses $n{=}200$ validation scenes per scenario that are disjoint from training ($10{,}800$ inferences per model). DPO+L2+SAM+Adv starts from DPO+L2+SAM and adds $162$ adversarial pairs to the $1500$ natural pairs ($\sim\!10\%$ adversarial). DPO+L2+SAM+Adv (scaled) uses $500$ adversarial pairs ($\sim\!25\%$ adversarial). In the last column and in the row of improved scripts, base denotes this starting model, not the base planner.}}
\label{tab:awsim_adv_train}
\resizebox{\ifdim\width>\linewidth\linewidth\else\width\fi}{!}{%
\begin{tabular}{lrrrr}
\toprule
Family ($\#$ scripts) & DPO+L2+SAM & \rev{$+$Adv ($162$ pairs)} & \rev{$+$Adv ($500$ pairs)} & \rev{$\Delta$ ($500$ vs.\ base)} \\
\midrule
cutin    ($42$)      & $18.51\%$ & $19.57\%$ & $\mathbf{17.07\%}$ & $-1.44$\,pp \\
cutout   ($\phantom{0}8$) & $1.00\%$  & $0.06\%$  & $\mathbf{0.00\%}$  & $-1.00$\,pp \\
decel    ($\phantom{0}4$) & $53.50\%$ & $54.75\%$ & $\mathbf{51.12\%}$ & $-2.38$\,pp \\
\midrule
peak decel p95 (m/s$^2$) & $1.20$ & $2.20$ & $\mathbf{4.04}$ & $+2.84$ \\
\# scripts improved vs.\ base & --- & $13/54$ & $\mathbf{31/54}$ & --- \\
natural-data comfort (n=500) & $0.00\%$ & $0.00\%$ & $0.00\%$ & $\pm 0.00$ \\
\bottomrule
\end{tabular}}
\end{table}

\begin{figure}[!htbp]
\centering
\includegraphics[width=0.96\linewidth]{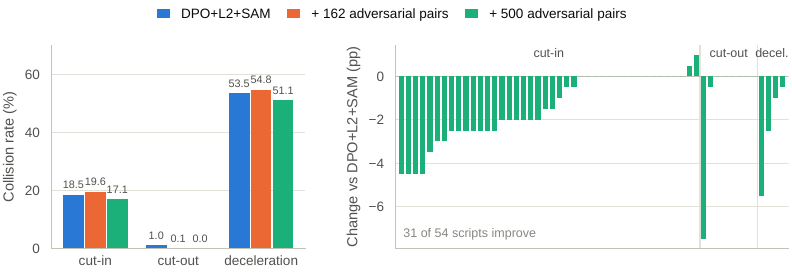}
\caption{\rev{Effect of training with adversarial pairs ($n{=}200$ disjoint validation scenes per scenario, $10{,}800$ inferences per model). Left: collision~\% by family for DPO+L2+SAM, DPO+L2+SAM+Adv ($162$ pairs) and DPO+L2+SAM+Adv (scaled, $500$ pairs). Right: per-scenario difference between DPO+L2+SAM+Adv (scaled) and DPO+L2+SAM for all $54$ scripts, sorted by family.}}
\label{fig:awsim_adv_train}
\end{figure}

\paragraph{\rev{Effect of the adversarial pool.}} \rev{With $500$ adversarial pairs (DPO+L2+SAM+Adv (scaled), $\sim\!25\%$ of the mixture), $31/54$ scripts improve. Cut-in collisions fall to $17.07\%$ from $18.51\%$ for DPO+L2+SAM, cut-out collisions fall to $0\%$, and the peak ego deceleration rises to $4.04$\,m/s$^2$ ($3.4{\times}$ the natural-data baseline). The deceleration family remains at $51.12\%$, because the chosen trajectory brakes at no more than $4$\,m/s$^2$ for comfort, which is too weak for an NPC braking at $1$\,G. An ablation that instead generates the chosen trajectory with an IDM lookahead~\citep{Treiber2000-dr} ($a_{\max}{=}1.5$\,m/s$^2$, $T_{\mathrm{hw}}{=}1.5$\,s, $s_0{=}2$\,m, $483$ pairs) performs worse on every family (cut-in $22.02\%$, peak deceleration $2.87$\,m/s$^2$), which indicates that the generator of chosen trajectories is not the bottleneck.}

\subsection{\rev{Relation to the Main Text}}

\rev{The V3 study and the adaptation methods of the main text share the base planner (adapter stack with shared rank $4$ and expert rank $8$, $\alpha{=}32$) and the ODD-Kashiwa site, but they address different questions. The main-text methods (\S\ref{sec:routing_fails}--\S\ref{sec:joint_replay_frontier}) optimise an L2 loss on target-region lanelets for transfer to new regions. The V3 study optimises the hybrid DPO+L2 loss on ODD-Kashiwa to close the comfort regression.}

\section{\rev{Trajectory Diagnostics on the Target-Region Pool}}
\label{app:joint_replay_safety}
\rev{The diagnostics in this section are proxies for comfort and corridor compliance, not closed-loop safety measurements. They are computed in open loop from a single deterministic DPM-Solver$++$ rollout per scene on the $24$-scene pool. Time-to-collision and collision rates are undefined on these scenes, because they contain no neighbouring agents.}

\begin{table}[h]
\centering\footnotesize
\caption{\rev{Trajectory diagnostics on the $24$-scene held-out target-region pool. The quantities are computed in open loop from a single deterministic DPM-Solver$++$ rollout per scene; they are proxies for comfort and corridor compliance, not closed-loop safety measurements. Jerk and lateral acceleration are unsmoothed finite differences of the raw $10$\,Hz plan, so only comparisons between models are meaningful; exit-\% is the share of rollouts that leave a straight $\pm 3$\,m band along the start lanelet. Of the $10$ Ishiodai scenes, $7$ are in the training set of both joint-replay checkpoints (\S\ref{sec:routing_fails}).}}
\label{tab:joint_replay_safety}
\setlength{\tabcolsep}{4pt}
\resizebox{\ifdim\width>\linewidth\linewidth\else\width\fi}{!}{%
\begin{tabular}{lrrrrr}
\toprule
configuration & lateral offset (m) & jerk-p95 & lat-acc-p95 & exit-\% & progress (m) \\
\midrule
shared module only (no expert)        & $24.89$ & $1323$ & $44.31$ & $\phantom{0}95.8$ & $164.3$ \\
static default (ODD-Nagoya)               & $16.04$ & $\phantom{0}970$ & $54.40$ & $100.0$ & $163.7$ \\
joint-replay (inherit-shared)  & $\phantom{0}6.95$ & $\mathbf{\phantom{0}719}$ & $52.25$ & $100.0$ & $\mathbf{186.9}$ \\
\textbf{joint-replay (from-scratch r=4)} & $\mathbf{\phantom{0}4.35}$ & $\phantom{0}848$ & $\mathbf{23.47}$ & $\mathbf{\phantom{0}50.0}$ & $178.1$ \\
\bottomrule
\end{tabular}}
\end{table}

\section{\rev{Held-Out Target-Region Topologies}}
\label{app:topologies}

\begin{figure}[t]
\centering
\includegraphics[width=0.97\linewidth]{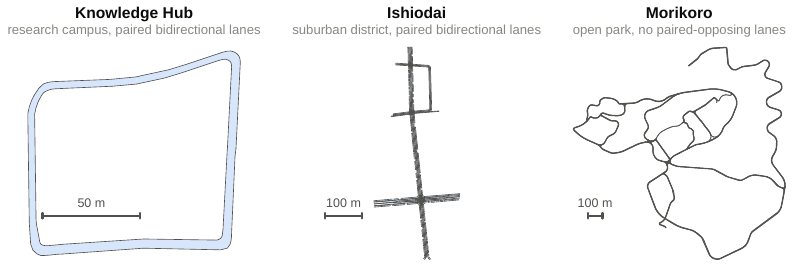}
\caption{\rev{Real-world topologies of the three held-out target-region maps. Every training ODD has paired opposing lanes. This structure is preserved on Knowledge Hub (research campus) and Ishiodai (suburban district) but absent on Morikoro (open park), which is the geometric source of the $31.9$\,m lateral offset of the static default analysed in \S\ref{sec:routing_fails}.}}
\label{fig:realvehicle}
\end{figure}

\section{\rev{Regions Where the Nearest Expert Is Not ODD-Nagoya}}
\label{app:switching}
\rev{On the three held-out maps of the main study, the geographically nearest training ODD is ODD-Nagoya, which is also the best fixed expert, so proximity and expert quality cannot be separated there. Table~\ref{tab:switching} therefore reports two further regions in which the static default selects a different expert. In both regions, the nearest expert is not the best fixed expert. The first region is a suburban bus-route map adjacent to ODD-Kashiwa. We exclude scenes whose start lanelet falls inside the bounding box of the ODD-Kashiwa training map ($398$ lanelets), which leaves $30$ scenes. Here the nearest expert (ODD-Kashiwa) is worse than ODD-Nagoya by $1.37$\,m (paired Wilcoxon $p{=}0.001$). The second region is a dense urban area nearest ODD-Hongo ($10$ scenes). Data from this area were used to train the shared module, because it is the ODD of a fifth expert that is not part of the four-expert adapter stack. This region therefore tests expert selection only, not adaptation to held-out data. Here the nearest expert (ODD-Hongo) is the weakest of the four and is $30.0$\,m worse than the best fixed expert (ODD-Kashiwa); with $10$ scenes, this difference is not significant ($p{=}0.084$).}

\begin{table}[h]
\centering\footnotesize\setlength{\tabcolsep}{4pt}
\caption{\rev{Mean maximum lateral offset (m) of each fixed expert in two regions where the static default does not select ODD-Nagoya. Bold: best fixed expert. The static default selects the nearest expert. $^\dagger$Data from this area were used to train the shared module.}}
\label{tab:switching}
\resizebox{\ifdim\width>\linewidth\linewidth\else\width\fi}{!}{%
\begin{tabular}{lcccccccc}
\toprule
 & & \multicolumn{4}{c}{\rev{Fixed expert (ODD-)}} & & & \\
\cmidrule(lr){3-6}
\rev{Region} & \rev{$n$} & \rev{Hongo} & \rev{Kashiwa} & \rev{Shalun} & \rev{Nagoya} & \rev{Shared only} & \rev{Nearest} & \rev{Best} \\
\midrule
\rev{Suburb near ODD-Kashiwa} & \rev{$30$} & \rev{$19.70$} & \rev{$12.70$} & \rev{$15.90$} & \rev{$\mathbf{11.33}$} & \rev{$15.59$} & \rev{Kashiwa} & \rev{Nagoya} \\
\rev{Urban area nearest ODD-Hongo$^\dagger$} & \rev{$10$} & \rev{$38.62$} & \rev{$\mathbf{8.61}$} & \rev{$13.97$} & \rev{$8.65$} & \rev{$14.60$} & \rev{Hongo} & \rev{Kashiwa} \\
\bottomrule
\end{tabular}}
\end{table}

\end{document}